\def\edo{early dropout}
\def\esd{early s.d.}

\def\ldo{late dropout}
\def\lsd{late s.d.}

\def\cdo{standard dropout}
\def\csd{standard s.d.}

\def\ndo{no dropout}
\def\nsd{no s.d.}

\def\convnext{ConvNeXt-F}
\def\swin{Swin-F}
\def\vit{ViT-T}
\def\mixer{Mixer-S}

\def\vitb{ViT-B}
\def\mixerb{Mixer-B}

\documentclass{article}

\usepackage{graphicx}
\usepackage{amsmath}
\usepackage{amssymb}
\usepackage{booktabs}
\usepackage{multirow}
\usepackage{makecell}
\usepackage{times}
\usepackage{microtype}
\usepackage{epsfig}
\usepackage[table,xcdraw,dvipsnames]{xcolor}
\usepackage{caption}
\usepackage{float}
\usepackage{placeins}
\usepackage{color, colortbl}
\usepackage{stfloats}
\usepackage{enumitem}
\usepackage{tabularx}
\usepackage{xstring}
\usepackage{multirow}
\usepackage{xspace}
\usepackage{url}
\usepackage{subcaption}
\usepackage{siunitx}
\usepackage[british, english, american]{babel}
\usepackage{graphicx, amsmath, amssymb, caption, subcaption, multirow, overpic, textpos}

\newcolumntype{x}[1]{>{\centering\arraybackslash}p{#1pt}}
\newcolumntype{y}[1]{>{\raggedright\arraybackslash}p{#1pt}}
\newcolumntype{z}[1]{>{\raggedleft\arraybackslash}p{#1pt}}

\definecolor{baselinecolor}{gray}{.95}
\newcommand{\gr}{\rowcolor[gray]{.95}}

\newcommand{\bb}[1]{\textbf{#1}}

\definecolor{green}{HTML}{009000} 
\definecolor{red}{HTML}{ea4335}  % red

\newcommand{\hlg}[1]{\textcolor{green}{#1}}
\newcommand{\hlr}[1]{\textcolor{red}{#1}}
\newcommand{\better}[1]{\hlg{$\uparrow\,$#1}}
\newcommand{\worse}[1]{\hlr{$\downarrow\,$#1}}

\newcommand{\betterinv}[1]{\hlg{$\downarrow\,$#1}}
\newcommand{\worseinv}[1]{\hlr{$\uparrow\,$#1}}

\newlength\savewidth\newcommand\shline{\noalign{\global\savewidth\arrayrulewidth
  \global\arrayrulewidth 1pt}\hline\noalign{\global\arrayrulewidth\savewidth}}
\newcommand{\tablestyle}[2]{\setlength{\tabcolsep}{#1}\renewcommand{\arraystretch}{#2}\centering\footnotesize}

\usepackage[pagebackref=false, breaklinks=true, letterpaper=true, colorlinks,
            citecolor=citecolor, linkcolor=linkcolor, bookmarks=false]{hyperref}

\usepackage[accepted]{icml2023}
\renewcommand{\paragraph}[1]{\vspace{0.0mm}\noindent\textbf{#1}}

\icmltitlerunning{Dropout Reduces Underfitting}

\begin{document}

\twocolumn[
\icmltitle{Dropout Reduces Underfitting}

\icmlsetsymbol{equal}{*}

\begin{icmlauthorlist}
\icmlauthor{Zhuang Liu}{equal,meta}
\icmlauthor{Zhiqiu Xu}{equal,cal}
\icmlauthor{Joseph Jin}{cal}
\icmlauthor{Zhiqiang Shen}{hku}
\icmlauthor{Trevor Darrell}{cal}
\end{icmlauthorlist}

\icmlaffiliation{meta}{FAIR, Meta AI}
\icmlaffiliation{cal}{UC Berkeley}
\icmlaffiliation{hku}{MBZUAI}

\icmlcorrespondingauthor{Zhuang Liu}{zhuangl@meta.com}
\icmlcorrespondingauthor{Zhiqiu Xu}{oscar.xzq@berkeley.edu}

\vskip 0.3in
]

\printAffiliationsAndNotice{\icmlEqualContribution} % otherwise use the standard text.

\begin{abstract}
Introduced by Hinton et al. in 2012, dropout has stood the test of time as a regularizer for preventing overfitting in neural networks.
In this study, we demonstrate that dropout can also mitigate \emph{underfitting} when used at the start of training.
During the early phase, we find dropout reduces the directional variance of gradients across mini-batches and helps align the mini-batch gradients with the entire dataset's gradient. This helps counteract the stochasticity of SGD and limit the influence of individual batches on model training.
Our findings lead us to a solution for improving performance in underfitting models - \emph{early dropout}: dropout is applied only during the initial phases of training, and turned off afterwards.
Models equipped with early dropout achieve \emph{lower} final training loss compared to their counterparts without dropout. 
Additionally, we explore a symmetric technique for regularizing overfitting models - \emph{late dropout}, where dropout is not used in the early iterations and is only activated later in training.
Experiments on ImageNet and various vision tasks demonstrate that our methods consistently improve generalization accuracy. 
Our results encourage more research on understanding regularization in deep learning and our methods can be useful tools for future neural network training, especially in the era of large data.
Code is available at \url{https://github.com/facebookresearch/dropout}.
\end{abstract}

\section{Introduction}
\label{sec:intro}
The year 2022 marks a full decade since AlexNet's pivotal ``ImageNet moment''~\cite{Krizhevsky2012}, which launched a new era in deep learning. 
It is no coincidence that dropout~\cite{Hinton2012} also celebrates its tenth birthday in 2022: AlexNet employed dropout to substantially reduce its overfitting, which played a critical role in its victory at the ILSVRC 2012 competition. 
Without the invention of dropout, the advancements we currently see in deep learning might have been delayed by years.

Dropout has since become widely adopted as a regularizer to mitigate overfitting in neural networks. It randomly deactivates each neuron with probability $p$, preventing different features from co-adapting with each other~\cite{Hinton2012,Srivastava2014}. After applying dropout, training loss typically increases, while test error decreases, narrowing the model's generalization gap.

Deep learning evolves at an incredible speed. Novel techniques and architectures are continuously introduced, applications expand, benchmarks shift, and even convolution can be gone~\cite{Dosovitskiy2021} -- but dropout has stayed.
It continues to function in the latest AI achievements, including AlphaFold's protein structure prediction~\cite{jumper2021highly}, and DALL-E 2's image generation~\cite{ramesh2022hierarchical}, demonstrating its versatility and effectiveness.

Despite the sustained popularity of dropout, its strength, represented by the drop rate $p$, has generally been decreasing over the years.
In the original dropout work~\cite{Hinton2012}, a default drop rate of 0.5 was used. However, lower drop rates, such as 0.1, have been frequently adopted in recent years. Examples include  training BERT~\cite{devlin2018bert} and Vision Transformers~\cite{Dosovitskiy2021}.

The primary driver for this trend is the exploding growth of available training data, making it increasingly difficult to overfit. 
In addition, advancements in data augmentation techniques~\cite{Zhang2018a,Cubuk2020} and algorithms for learning with unlabeled or weakly-labeled data~\cite{Brown2020,Radford2021,he2021masked} have provided even more data to train on than the model can fit to. As a result, we may soon be confronting more problems with \emph{underfitting} instead of overfitting.

\begin{figure}[t]\centering
\vspace{-.6em}
\includegraphics[width=0.98\linewidth]{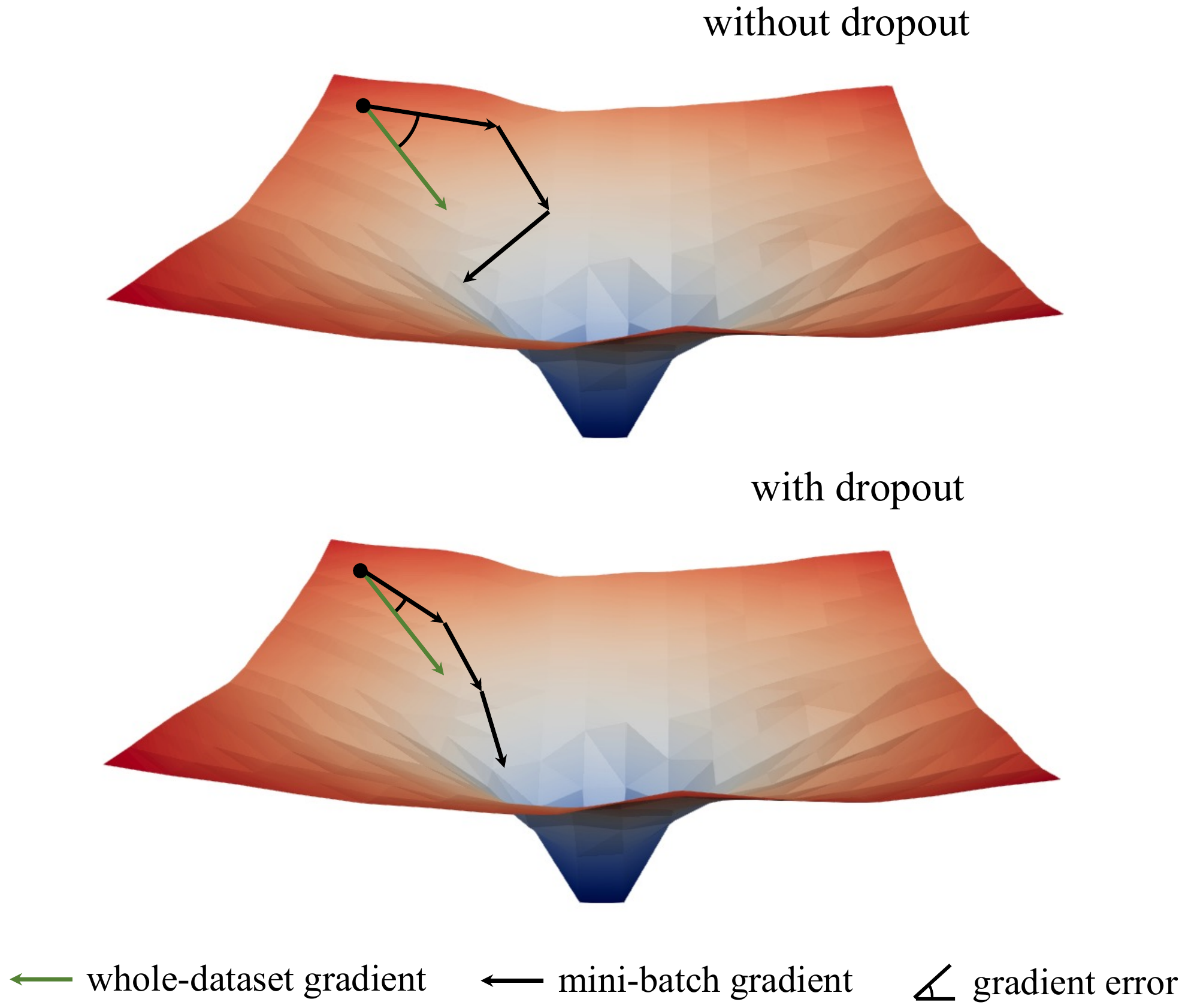}\\
\vspace{-1.1em}
\caption{\textbf{Dropout in early training} helps the model produce mini-batch gradient directions that are more consistent and aligned with the overall gradient of the entire dataset.}
\vspace{-1.9em}
\label{fig:fig1}
\end{figure}

Would dropout lose its relevance should such a situation arise? In this study, we demonstrate an alternative use of dropout for tackling underfitting. We begin our investigation into dropout training dynamics by making an intriguing observation on gradient norms, which then leads us to a key empirical finding: during the initial stages of training, dropout reduces gradient variance across mini-batches and allows the model to update in more consistent directions. These directions are also more aligned with the entire dataset's gradient direction (Figure~\ref{fig:fig1}). Consequently, the model can optimize the training loss more effectively with respect to the whole training set, rather than being swayed by individual mini-batches. 
In other words, dropout counteracts SGD and prevents excessive regularization due to randomness in sampling mini-batches during early training.

\begin{figure}[t]\centering
\vspace{-.3em}
\includegraphics[width=0.98\linewidth]{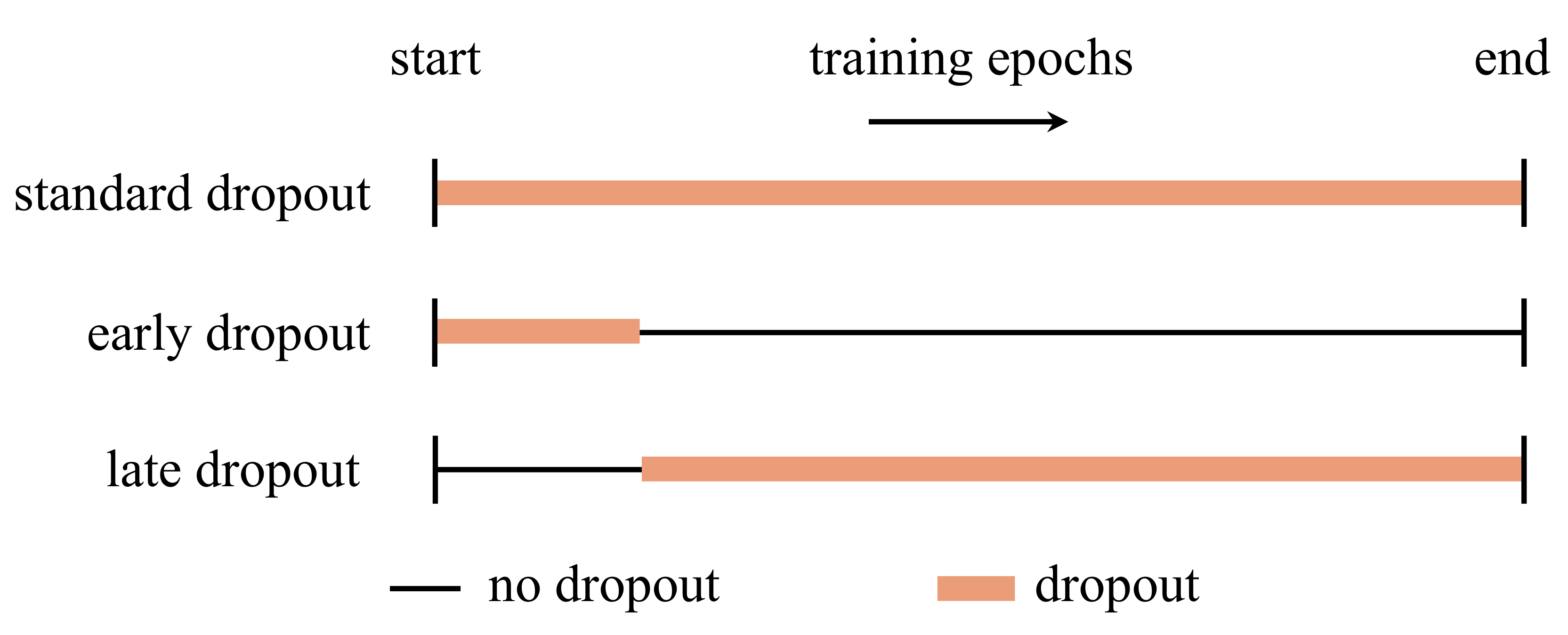}\\
% \vspace{-.5em}
\caption{\textbf{Standard, early and late dropout.} We propose early and late dropout. Early dropout helps underfitting models fit the data better and achieve lower training loss. Late dropout helps improve the generalization performance of overfitting models. 
}
\label{fig:dropout}
\vspace{-.5em}
\end{figure}

Based on this insight, we introduce \emph{\edo{}} -- dropout is only used during early training -- to help underfitting models fit better.
Early dropout \emph{lowers} the final training loss compared to no dropout and standard dropout.
Conversely, for models that already use standard dropout, we propose to remove dropout during earlier training epochs to mitigate overfitting. We refer to this approach as \emph{\ldo{}} and demonstrate that it improves generalization accuracy for large models. Figure~\ref{fig:dropout} provides a comparison of standard dropout, \edo{}, and \ldo{}.

We evaluate early and late dropout using different models on image classification and downstream tasks. Our methods consistently yield better results than both standard dropout and no dropout.
We hope our findings can offer novel insights into dropout and overfitting, and motivate further research in developing neural network regularizers.

\section{Revisiting Overfitting vs. Underfitting}
\label{sec:background}

\paragraph{Overfitting.}
Overfitting occurs when a model is trained to fit the training data excessively well but generalizes poorly to unseen data. The model's capacity and the dataset scale are among the most critical factors in determining overfitting, along with other factors such as training length. Larger models and smaller datasets tend to lead to more overfitting.

\begin{figure}[h]\centering
\includegraphics[width=.98\linewidth]{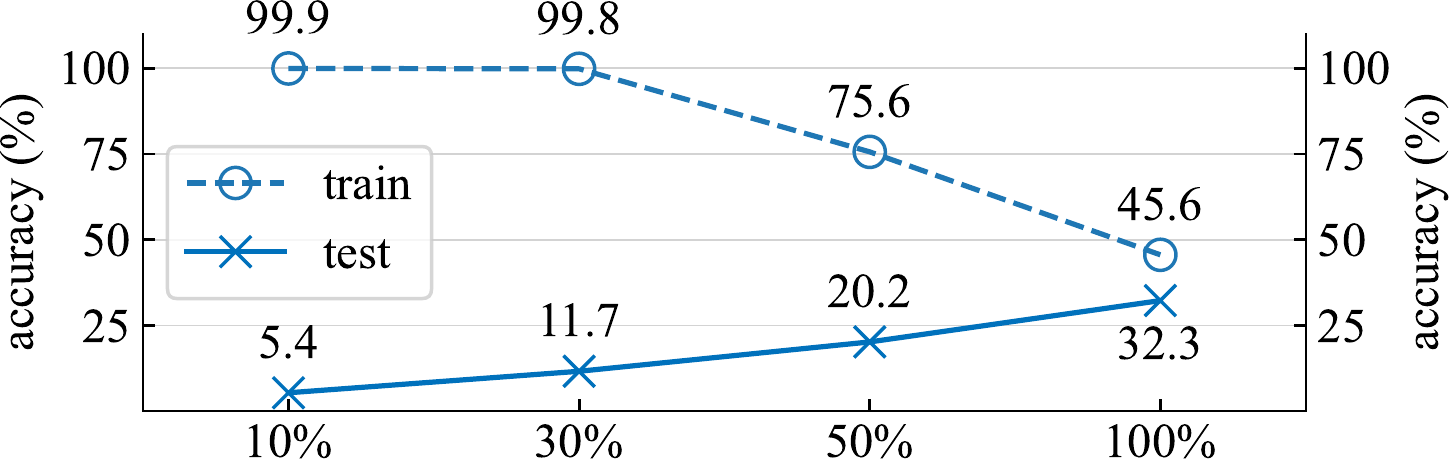}\\
\scriptsize amount of ImageNet train data (ViT-T) \\
\vspace{1em}
\includegraphics[width=.98\linewidth]{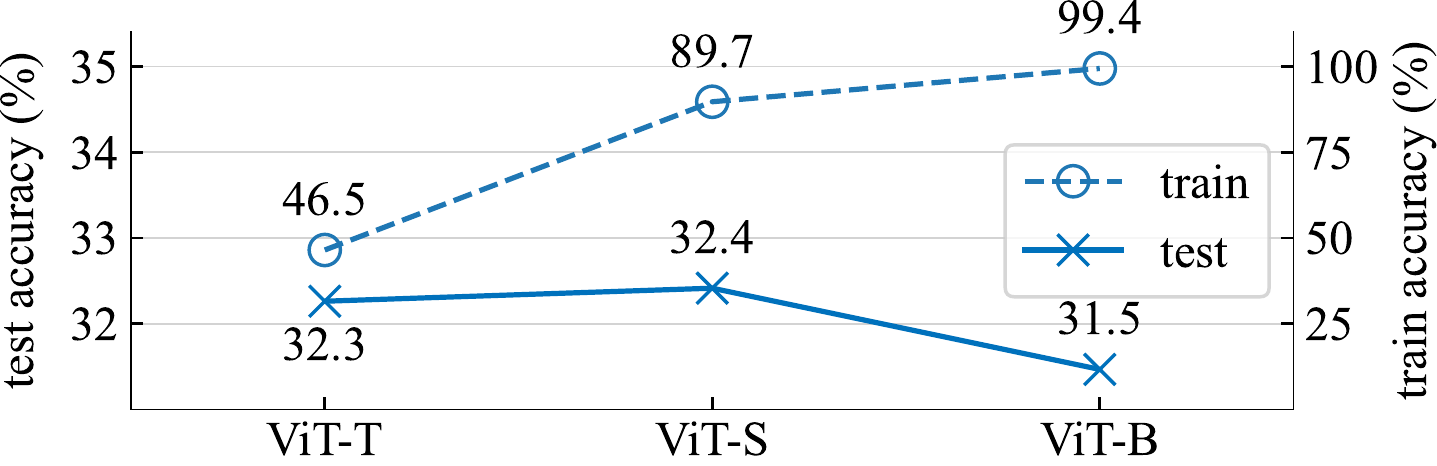}\\
\caption{\textbf{Overfitting} can occur when either the amount of data decreases (top) or the capacity of the model increases (bottom). 
}
\label{fig:mot_1}
\end{figure}

We conduct several simple experiments to clearly illustrate this trend. First, when the model remains the same, but we use less data, the gap between training accuracy and test accuracy increases, leading to overfitting. Figure~\ref{fig:mot_1} (top) demonstrates this trend with ViT-Tiny/32 results trained on various amounts of ImageNet data.
Second, when the model capacity increases while keeping the dataset size constant, the gap also widens. Figure~\ref{fig:mot_1} (bottom) illustrates this with ViT-Tiny (T), Small (S), and Base (B)/32 models trained on the same 100\% ImageNet data. We train all models with a fixed 4,000 iterations without data augmentations.

\paragraph{Dropout.}
We briefly review the dropout method. 
At each training iteration, a dropout layer randomly sets each neuron to zero with a certain probability for its input tensor. During inference, all neurons are active but are scaled by a coefficient to maintain the same overall scale as in training.
As each sample is trained by a different sub-network, dropout can be seen as an implicit ensemble of exponentially many models. 
It is a fundamental building block of deep learning and has been used to prevent overfitting in various of neural architectures and applications~\cite{Vaswani2017,devlin2018bert,ramesh2022hierarchical}. 

\paragraph{Stochastic depth.} 
Various efforts have been made to design dropout variants~\cite{Wan2013,He2014,ghiasi2018dropblock}. 
In this work, we also consider a dropout variant called stochastic depth~\cite{Huang2016deep} (s.d. for short), which is designed for regularizing residual networks~\cite{He2016}. For each sample or mini-batch, the network randomly selects a subset of residual blocks to skip, making the model shallower and thus earning its name ``stochastic depth''. 
It is commonly seen in modern vision networks, including DeiT~\cite{Touvron2020}, ConvNeXt~\cite{liu2022convnet} and MLP-Mixer~\cite{tolstikhin2021mlp}. 
Several recent models~\cite{steiner2021train,tolstikhin2021mlp} use s.d.  together with dropout. 
Since s.d. can be viewed as specialized dropout at the residual block level, the term ``dropout'' that we use later could also encompass s.d., depending on the context.

\paragraph{Drop rate.}
The probability of setting a neuron to zero in dropout is referred to as the drop rate $p$, a hugely influential hyper-parameter. 
As an example, in Swin Transformers and ConvNeXts, the only training hyper-parameter that varies with the model size is the stochastic depth drop rate.

\begin{figure}[h]\centering
\includegraphics[width=.98\linewidth]{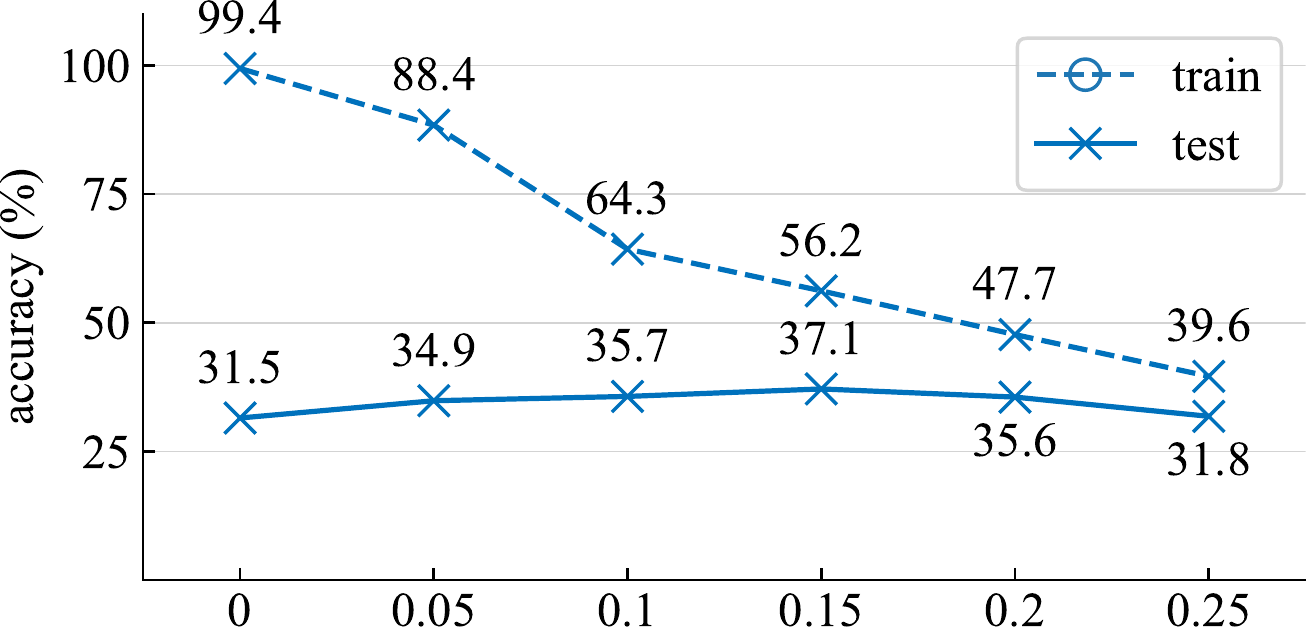}\\
\hspace{1em} \scriptsize drop rate (ViT-B)\\
\vspace{-0.7em}
\caption{\textbf{Drop rate influence}. The training accuracy decreases as the drop rate increases. However, there is an optimal drop rate ($p=0.15$ in this case) that maximizes the test accuracy.
}
\label{fig:mot_1.5}
\end{figure}

We apply dropout to regularize the ViT-B model and experiment with different drop rates. As shown in Figure~\ref{fig:mot_1.5}, setting the drop rate too low does not effectively prevent overfitting, whereas setting it too high results in over-regularization and decreased test accuracy. In this case, the optimal drop rate for achieving the highest test accuracy is 0.15.

\begin{figure}[h]\centering
\vspace{1.3em}
\includegraphics[width=.98\linewidth]{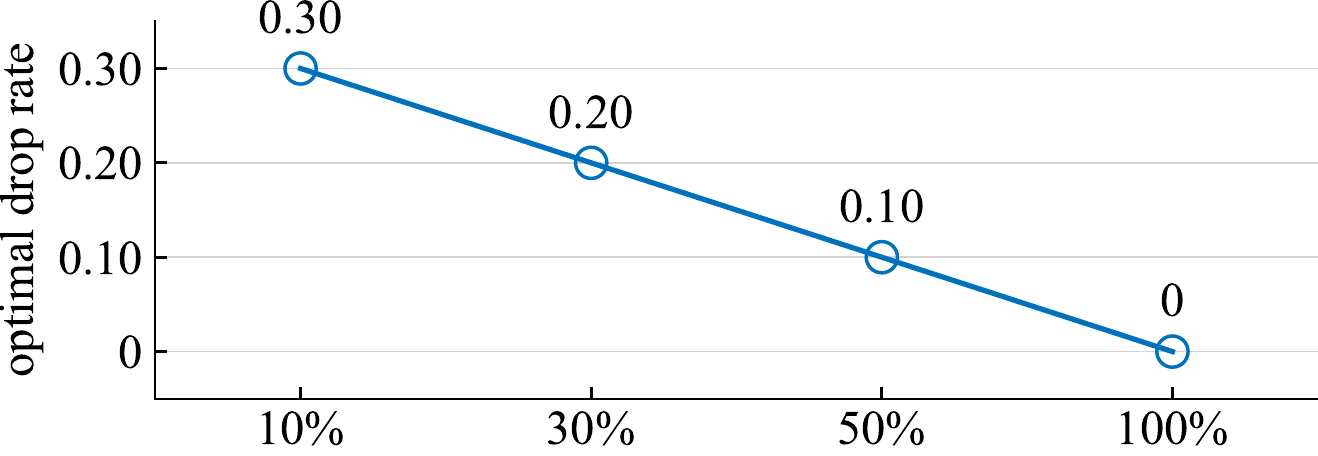}\\
\scriptsize \qquad amount of ImageNet data (ViT-T)\\
\includegraphics[width=.98\linewidth]{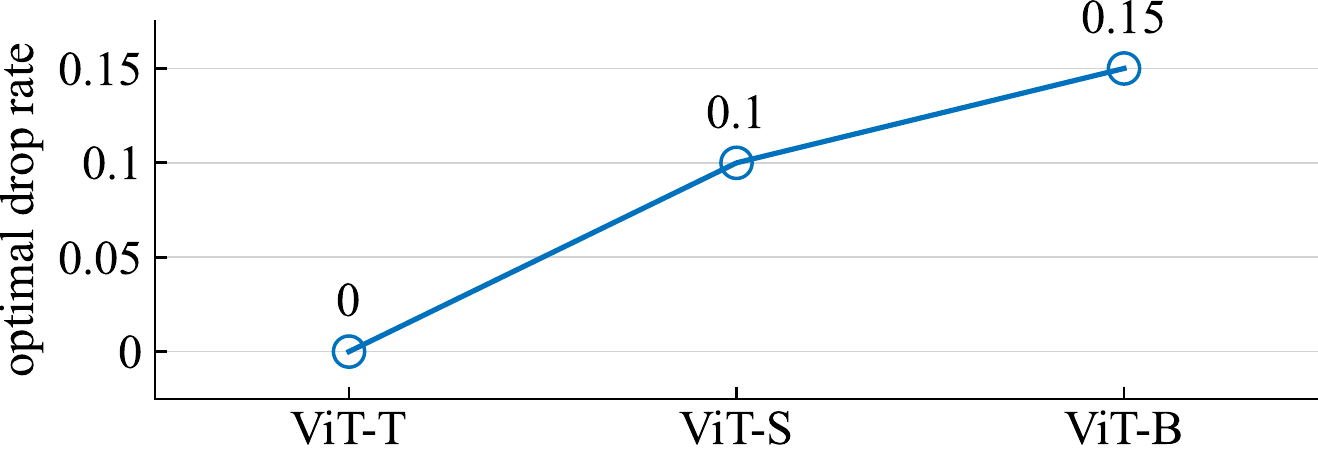}\\
\caption{\textbf{Optimal drop rate}. Training with a larger dataset (top) or using a smaller model (bottom) both result in a lower optimal drop rate, which may even reach 0 in some cases.
}
\label{fig:mot_2}
\end{figure}

Different model architectures use different drop rates, and the selection of optimal drop rate $p$ heavily depends on the network model size and the dataset size. In Figure~\ref{fig:mot_2}, we plot the best dropout rate for model and data settings from Figure~\ref{fig:mot_1}. We perform a hyper-parameter sweep for drop rate at intervals of 0.05 for each setting. From Figure~\ref{fig:mot_2}, we observe that when the data is large enough, or when the model is small enough, the best drop rate $p$ is 0, indicating that using dropout may not be necessary and could harm the model's generalization accuracy by underfitting the data.

\paragraph{Underfitting.} In the literature, the drop rate used for dropout has generally decreased over the years. Earlier models such as VGG~\cite{Simonyan2015} and GoogleNet~\cite{Szegedy2015} use 0.5 or higher drop rates; ViTs~\cite{Dosovitskiy2021} use a moderate rate of 0.1 on ImageNet and do not use dropout when pre-training on the much larger JFT-300M dataset; recent language-supervised or self-supervised vision models~\cite{Radford2021,he2021masked} do not use dropout. This trend is likely due to the increasing size of datasets. The model does not overfit very easily to immense data. 

With the rapidly growing amount of data being generated and distributed globally, it is possible that the scale of the available data may soon outpace the capacities of the models we train. While data is generated at a speed of quintillion bytes per day, models still need to be stored and run on finite physical devices such as servers, data centers, or mobile phones. Given such a contrast, future models may have more trouble fitting data properly rather than overfitting too severely. As our experiments above demonstrate, in such settings, standard dropout may not help generalization as a regularizer. Instead, we need tools to help models fit vast amounts of data better and reduce \emph{underfitting}.

\section{How Dropout Can Reduce Underfitting}
\label{sec:how}

In this study, we explore whether dropout can be used as a tool to reduce underfitting. To this end, we conduct a detailed  
analysis on the training dynamics of dropout using our proposed tools and metrics. We compare two \vit{}/16 training processes on ImageNet~\cite{Deng2009}: one without dropout as the baseline, and the other with a 0.1 dropout rate throughout training.

\begin{figure}[h]
  \centering
  \includegraphics[width=.49\linewidth]{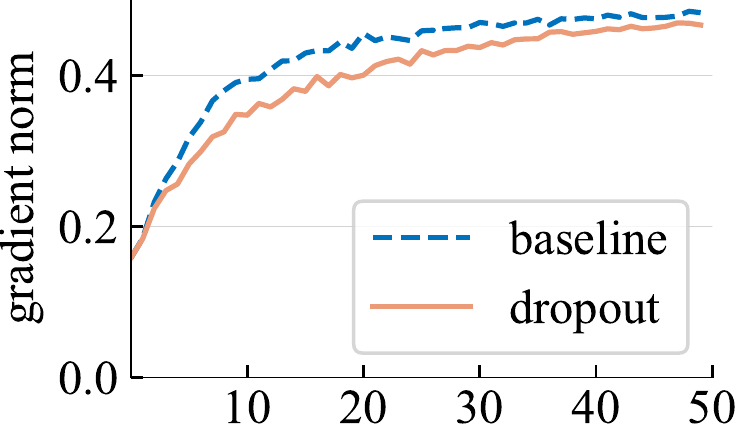}
  \hspace{-.3em}
  \includegraphics[width=.49\linewidth]{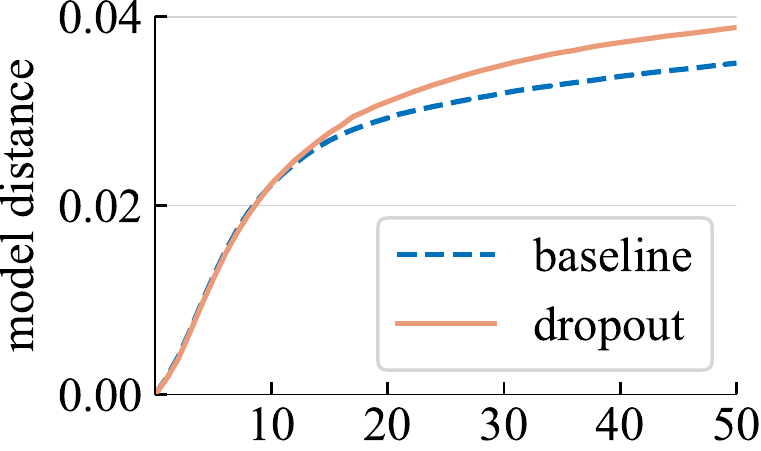}
  \vspace{-1em}
  \scriptsize \,\,\,\,\,\,epochs\\
  \vspace{.5em}
  \caption{\textbf{Gradient norm} (left) and \textbf{model distance} (right). The model with dropout has smaller gradient magnitudes, but it moves a greater distance in the parameter space.}
  \label{fig:grad_norm}
\end{figure}

\paragraph{Gradient norm.} 
We begin our analysis by investigating the impact of dropout on the strength of gradients $g$, measured by their $L_2$ norm $||g||_2$. For the dropout model, we measure the entire model's gradient, even though a subset of weights may have been deactivated due to dropout. As shown in Figure~\ref{fig:grad_norm} (left), the dropout model produces gradients with smaller norms, indicating that it takes smaller steps at each gradient update.

\paragraph{Model distance.} Since the gradient steps are smaller, we expect the dropout model to travel a smaller distance from its initial point than the baseline model. 
To measure the distance between the two models, we use the $L_2$ norm, represented by $||W_1 - W_2||_2$, where $W_i$ denotes the parameters of each model.
In Figure~\ref{fig:grad_norm} (right), we plot each model's distance from its random initialization.
However, to our surprise, the dropout model actually moved by a \emph{larger} distance than the baseline model, contrary to what we initially anticipated based on the gradient norms.

Let us imagine two people walking. One walks with large strides while the other walks with small strides. Despite this, the person with smaller strides covers a greater distance from their starting point over the same time period. Why? This may be because the person is walking in a more \emph{consistent} direction, whereas the person with larger strides may be taking random, meandering steps and not making much progress in any one particular direction.

\begin{figure}[h]\centering
\includegraphics[width=.98\linewidth]{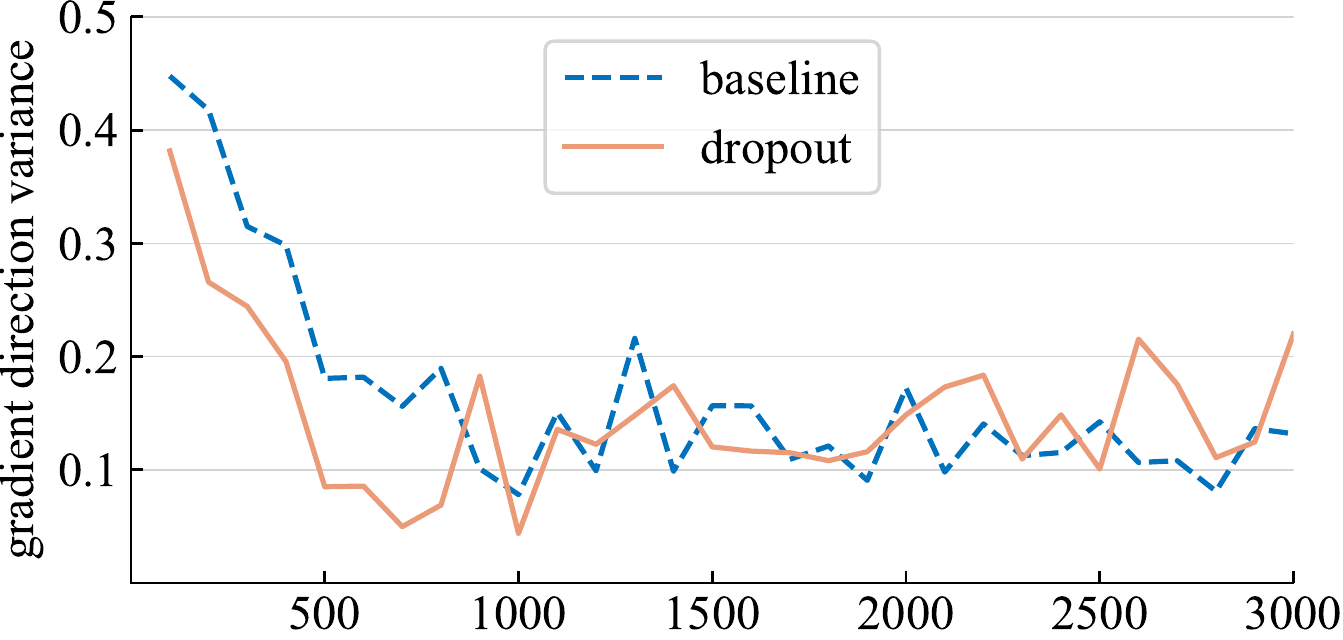}\\
\scriptsize iterations\\
\vspace{-0.7em}
\caption{\textbf{Gradient direction variance}. The model with dropout produces more consistent mini-batch gradients during the initial phase of training, up to approximately 1000 iterations.}
\label{fig:analysis_gradvar}
\end{figure}

\paragraph{Gradient direction variance.} We hypothesize the same for our two models: the dropout model is producing more consistent gradient directions across mini-batches. To test this, we collect a set of mini-batch gradients $G$ by training a model checkpoint on randomly selected batches. 
We propose to measure the gradient direction variance (GDV) by computing the average pairwise cosine distance: $$ GDV = \frac{2}{|G| \cdot (|G| - 1)}\sum_{g_i, g_j \in G, i \neq j} \underbrace{\frac{1}{2}(1 - \frac{<g_i, g_j>}{||g_i||_2 \cdot ||g_j||_2})}_{\text{cosine distance}}$$
As seen in Figure~\ref{fig:analysis_gradvar}, the comparison of variance supports our hypothesis. Up to a certain iteration (approximately 1000), the dropout model exhibits a lower gradient variance and moves in a more consistent direction.

Notably, prior work also studied the measure of gradient variances~\cite{jastrzebskibreak2020} or proposed methods to reduce gradient variance~\cite{johnson2013accelerating,balles2018dissecting,zhang2019lookahead,kavis2022adaptive} for optimization algorithms. Our metric is different in that only the gradient directions matter and each gradient equally contributes to the whole measurement.

\paragraph{Gradient direction error. }
However, the question remains -- what should be the correct direction to take?
To fit the training data, the underlying objective is to minimize the loss on the entire training set, not just on any single mini-batch. We compute the gradient for a given model on the whole training set, where dropout is set to inference mode to capture the full model's gradient. Then, we evaluate how far the actual mini-batch gradient $g_{step}$ is from this whole-dataset ``ground-truth'' gradient $\hat{g}$. We define the average cosine distance from all $g_{step} \in G$ to $\hat{g}$ as the gradient direction ``error'' (GDE):
$$ GDE = \frac{1}{|G|} \sum_{g_{step} \in G} \underbrace{\frac{1}{2}(1 - \frac{<g_{step}, \hat{g}>}{||g_{step}||_2 \cdot ||\hat{g}||_2})}_{\text{cosine distance}}$$

\begin{figure}[h]\centering
\includegraphics[width=.98\linewidth]{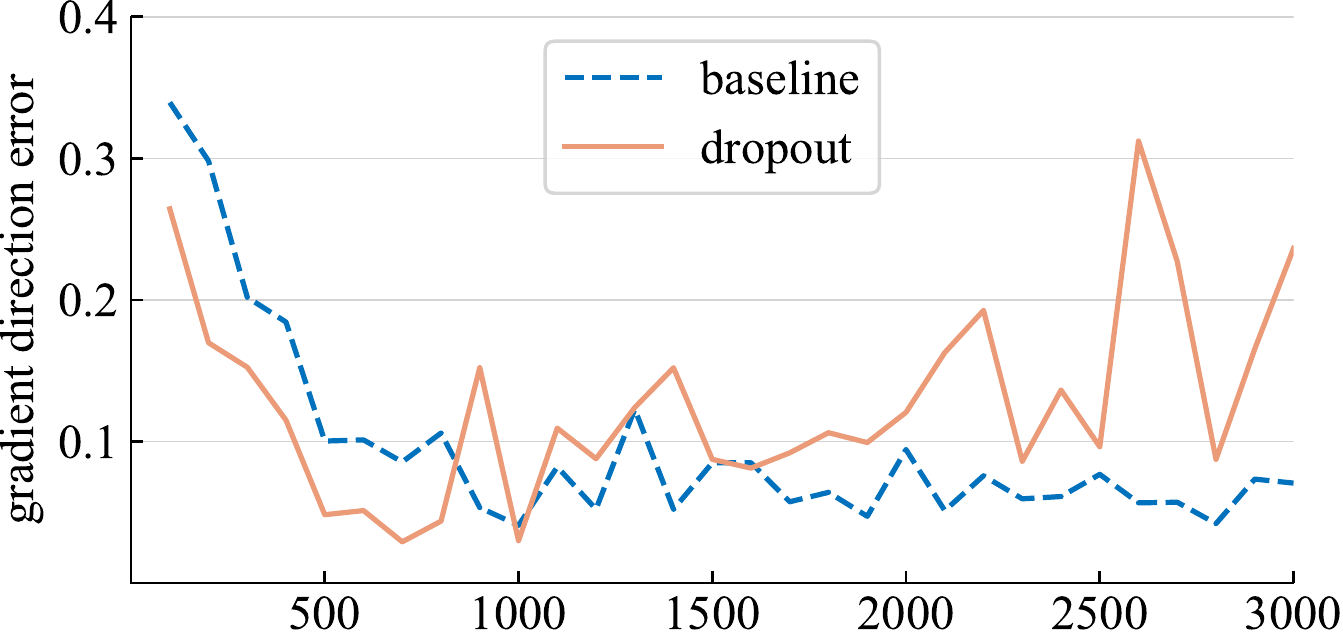}\\
\scriptsize iterations\\
\vspace{-0.7em}
\caption{\textbf{Gradient direction error}. Dropout leads to mini-batch gradients that are more aligned with the gradient of the entire dataset at the beginning of training.
}
\label{fig:analysis_graderr}
\end{figure}

We calculate this error term and plot it in Figure~\ref{fig:analysis_graderr}. At the beginning of training, the dropout model's mini-batch gradients have smaller deviations from the whole-dataset gradient, indicating that it is moving in a more desirable direction for optimizing the total training loss (as illustrated in Figure~\ref{fig:fig1}).
After approximately 1000 iterations, however, the dropout model produces gradients that are farther away. This could be the turning point where dropout transitions from reducing underfitting to reducing overfitting.

The experiments detailed above employ the ViT optimized with AdamW~\cite{Loshchilov2019}. We explore whether this observation remains consistent with other optimizers and architectures. To quantify the impact of gradient direction error (GDE) reduction, we measure the area under the curve (AUC) in the GDE vs. iteration plot (Figure~\ref{fig:analysis_graderr}) over the first 1500 iterations. This calculation represents the average GDE during this period, with a larger AUC value indicating higher GDE in initial training. We present the results in Table \ref{tab:analysis-graderr}. The reduction in gradient error is also observable with other optimizers and architectures, such as (momentum) SGD and Swin Transformer.

\begin{table}[h]
\centering
% \small
\footnotesize
\addtolength{\tabcolsep}{-5pt}
\def\arraystretch{1.3}% 
\vspace{-.4em}
\begin{tabular*}{0.48\textwidth}{@{\extracolsep{\fill}}l@{\hspace{-.3cm}}cccc}
   model &  optimizer & GDE & change  \\
\Xhline{1.0pt}
\vit{} (\ndo)  & AdamW      & 156.6 & -                 \\        \vit{} (\cdo) & AdamW & 135.3 & \betterinv{13.60\%}   \\
\Xhline{0.1pt}
\vit{} (\ndo)  & SGD      & 141.9 & -                 \\          \vit{} (\cdo) & SGD & 128.7 & \betterinv{9.30\%}   \\
\Xhline{0.1pt}
\vit{} (\ndo)  & momentum SGD      & 133.4 & -                 \\
\vit{} (\cdo) & momentum SGD & 124.5 & \betterinv{6.67\%}   \\
\Xhline{0.1pt}
\swin{} (\ndo)  & AdamW      & 718.4 & -                 \\       \swin{} (\cdo) & AdamW & 593.3 & \betterinv{17.41\%}   \\
   \swin{} (\csd) & AdamW & 583.8 & \betterinv{18.73\%}   \\
\Xhline{0.1pt}
\convnext{} (\nsd)  & AdamW      & 69.5 & -                 \\   \convnext{} (\csd) & AdamW & 64.2 & \betterinv{7.62\%}   \\
\Xhline{0.1pt}
\end{tabular*}
\vspace{-.3em}
\caption{\textbf{GDE reduction on different models and optimizers.}   We observe consistent GDE reduction for different models and optimizers at early training.}
\vspace{-.5em}
\label{tab:analysis-graderr}
\end{table} 

\paragraph{Bias and variance for gradient estimation.} This analysis at early training can be viewed through the lens of the bias-variance tradeoff. For no-dropout models, an SGD mini-batch provides an unbiased estimate of the whole-dataset gradient because the expectation of the mini-batch gradient is equal to the whole-dataset gradient, and each mini-batch runs through the same network. 
However, with dropout, the estimate becomes biased, as the mini-batch gradients are generated by different sub-networks, whose expected gradient may not match the full network's gradient. Nevertheless, the gradient variance is significantly reduced in our empirical observation, leading to a reduction in gradient error.
Intuitively, this reduction in variance and error helps prevent the model from overfitting to specific batches, especially during the early stages of training when the model is undergoing significant changes.

\section{Approach}
\label{sec:method}
From the analysis above, we know that using dropout early can potentially improve the model's ability to fit the training data. Based on this observation, we present our approaches.

\paragraph{Underfitting and overfitting regimes.} Whether it is desirable to fit the training data better depends on whether the model is in an underfitting or overfitting regime, which can be difficult to define precisely. In this work, we use the following criterion and find it is effective for our purpose: if a model generalizes better with standard dropout, we consider it to be in an overfitting regime; if the model performs better without dropout, we consider it to be in an underfitting regime. The regime a model is in depends not only on the model architecture but also on the dataset used and other training parameters. 

\paragraph{Early dropout.} In their default settings, models at underfitting regimes do not use dropout. To improve their ability to fit the training data, we propose \emph{\edo{}}: \emph{using dropout before a certain iteration, and then disabling it for the rest of training}. Our experiments show that early dropout reduces final training loss and improves accuracy.

\paragraph{Late dropout.} Overfitting models already have standard dropout included in their training settings. During the early stages of training, dropout may cause overfitting unintentionally, which is not desirable. To reduce overfitting, we propose \emph{\ldo{}}: \emph{not using dropout before a certain iteration, and then using it for the rest of training.} This is a symmetric approach to early dropout.

\paragraph{Hyper-parameters.} 
Our methods are straightforward both in concept and implementation, illustrated in Figure~\ref{fig:dropout}. 
They require two hyper-parameters: 1) the number of epochs to wait before turning dropout on or off. Our results show that this choice can be robust enough to vary from 1\% to 50\% of the total epochs. 2) The drop rate $p$, which is similar to the standard dropout rate and is also moderately robust.

\section{Experiments}
\label{sec:imagenet}

We conduct empirical evaluations on ImageNet-1K classification with 1,000 classes and 1.2M training images~\cite{Deng2009} and report top-1 validation accuracy.

\subsection{Early Dropout}

\begin{table}[t]
\centering
\small
\addtolength{\tabcolsep}{-3.pt}
\def\arraystretch{1.09}% 
\vspace{-.8em}
\begin{tabular}{lcccc}
   model &  top-1 acc. & change & train loss & change  \\
\Xhline{1.0pt}
\multicolumn{5}{c}{\scriptsize{results with basic recipe}}   \\
\vit{}      & 73.9  & -            & 3.443 & -                 \\
+ \cdo{}    & 67.9  & \worse{6.0}  & 3.885 & \worseinv{0.442}  \\
+ \csd{}    & 72.6  & \worse{1.3}  & 3.681 & \worseinv{0.238}  \\
\gr+ \edo{} & \bb{74.3}  & \better{0.4} & 3.394 & \betterinv{0.049} \\
\gr+ \esd{} & \bb{74.4} & \better{0.5} & 3.435 & \betterinv{0.008} \\

\hline
\mixer{}$^*$& 68.7  & -            & - & - \\
\mixer{}    & 71.0  & -            & 3.635 & -                 \\
+ \cdo{}    & 67.1  & \worse{3.9}  & 4.058 & \worseinv{0.423}  \\
+ \csd{}    & 70.5  & \worse{0.5}  & 3.813 & \worseinv{0.178}  \\
\gr+ \edo{} & \bb{71.3}  & \better{0.3} & 3.591 & \betterinv{0.044} \\
\gr+ \esd{} & \bb{71.7}  & \better{0.7} & 3.552 & \betterinv{0.083} \\

\hline
\convnext{} & 76.1  & -            & 3.472 & -                 \\
+ \csd{}    & 75.5  & \worse{0.6}  & 3.647 & \worseinv{0.175}  \\
\gr+ \esd{} & \bb{76.3}  & \better{0.2} & 3.443 & \betterinv{0.029} \\

\hline
\swin{}     & 74.3  & -            & 3.411 & -                 \\
+ \cdo{}    & 71.6  & \worse{2.7}  & 3.717 & \worseinv{0.306}  \\
+ \csd{}    & 73.7  & \worse{0.6}  & 3.644 & \worseinv{0.233}  \\
\gr+ \edo{} & \bb{74.7}  & \better{0.4} & 3.378 & \betterinv{0.033} \\
\gr+ \esd{} & \bb{75.2}  & \better{0.9} & 3.353 & \betterinv{0.058} \\
\hline

\multicolumn{5}{c}{\scriptsize{results with improved recipe}}  \\  
\vit{}$^\dagger$ & 72.8  & -            & - & -                 \\
\vit{}$^\ddagger$ & 75.5  & -            & - & -                 \\
\vit{}      & 76.3  & -            & 3.033 & -                 \\
+ \cdo{}    & 71.5  & \worse{4.8}  & 3.437 & \worseinv{0.404}  \\
+ \csd{}    & 75.6  & \worse{0.7}  & 3.243 & \worseinv{0.210}  \\
\gr+ \edo{} & \bb{76.7}  & \better{0.4} & 2.991 & \betterinv{0.042} \\
\gr+ \esd{} & \bb{76.7}  & \better{0.4} & 3.022 & \betterinv{0.011} \\

\hline
\convnext{}$^\ddagger$ & 77.5  & -            & - & -                 \\
\convnext{} & 77.5  & -            & 3.011 & -                 \\
+ \csd{}    & 77.4  & \worse{0.1}  & 3.177 & \worseinv{0.166}  \\
\gr+ \esd{} & \bb{77.7}  & \better{0.2} & 2.990 & \betterinv{0.021} \\

\hline
\swin{}     & 76.1  & -            & 2.989 & -                 \\
+ \cdo{}    & 73.5  & \worse{2.6}  & 3.305 & \worseinv{0.316}  \\
+ \csd{}    & 75.6  & \worse{0.5}  & 3.241 & \worseinv{0.252}  \\
\gr+ \edo{} & \bb{76.6}  & \better{0.5} & 2.966 & \betterinv{0.023} \\
\gr+ \esd{} & \bb{76.6}  & \better{0.5} & 2.958 & \betterinv{0.031} \\
\hline
\end{tabular}
\vspace{-0.7em}
\scriptsize
\caption{\textbf{Classification accuracy on ImageNet-1K.}  Early dropout or stochastic depth (s.d.) lowers training loss and improves test accuracy for underfitting models, while standard ones hurt both. Literature baselines: $*$ \citet{tolstikhin2021mlp}, $\dagger$\citet{Touvron2020}, $\ddagger$\citet{rw2019timm}.}
\vspace{-2.3em}
\label{tab:main-early}
\end{table}

\paragraph{Settings.} To evaluate \edo{}, we choose small models at underfitting regimes on ImageNet-1K, including \vit{}/16~\cite{Touvron2020}, \mixer{}/32~\cite{tolstikhin2021mlp}, ConvNeXt-Femto (F)~\cite{rw2019timm}, and a \swin{}~\cite{Liu2021swin} of similar size to \convnext{}. These models have 5-20M parameters and are relatively small for ImageNet-1K. We conduct separate evaluations for dropout and stochastic depth (s.d.), i.e., only one is used in each experiment. We use the training recipe from ConvNeXt~\cite{liu2022convnet} as our basic recipe. The drop rates are selected from ${0.1, 0.2, 0.3}$ for dropout and ${0.3, 0.5, 0.7}$ for s.d.
Each result is an average with 3 seeds, and the average standard deviation is 0.142\%. The usage of dropout does not affect training time noticeably. See Appendix for more details on the experimental setup and standard deviation results.

\paragraph{Results.}
Table~\ref{tab:main-early} (top) presents the results. Early dropout consistently improves the test accuracy, and also \emph{decreases} the training loss, indicating dropout at an early stage helps the model fit the data better. The results are compared to standard dropout and s.d. using a drop rate of 0.1, which both have a negative impact on the models.

Additionally, we double the training epochs and reduce mixup~\cite{Zhang2018a} and cutmix~\cite{Yun2019} strength to arrive at an improved recipe for these small models. Table~\ref{tab:main-early} (bottom) shows the results. The baselines now achieve much-improved accuracy, sometimes surpassing previous literature results by a large margin. Nevertheless, \edo{} still provides a further boost in accuracy.

\subsection{Analysis}

We carry out ablation studies to understand the characteristics of \edo{}. Our default setting is \vit{} training with \edo{} using the improved recipe.

\begin{figure}[h]\centering
% \hspace{-1em}
\includegraphics[width=.98\linewidth]{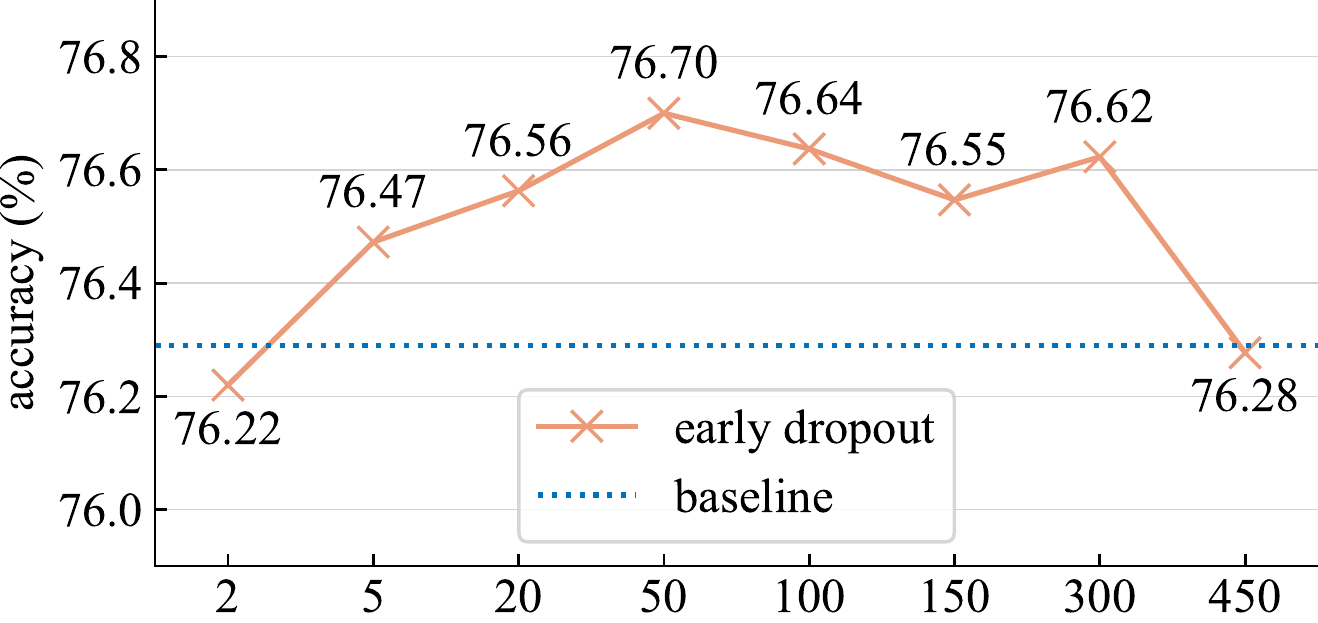}\\
\scriptsize \,\,\,\, epochs \\
\vspace{-.8em}
\caption{\textbf{Early dropout epochs}. Early dropout is effective with a wide range of dropout epochs.
}
\vspace{-.5em}
\label{fig:abl_epoch}
\end{figure}

\paragraph{Dropout epochs.}
We investigate the impact of the number of epochs for early dropout. By default, we use 50 epochs. We vary the number of early dropout epochs and observe its effect on the final accuracy. The results, shown in Figure~\ref{fig:abl_epoch}, are based on the average of 3 runs with different random seeds. The results indicate that the favorable range of epochs for both early dropout is quite broad, ranging from as few as 5 epochs to as many as 300, out of a total of 600 epochs. 
This robustness makes early dropout easy to adopt in practical settings.

\begin{figure}[h]\centering
\includegraphics[width=.98\linewidth]{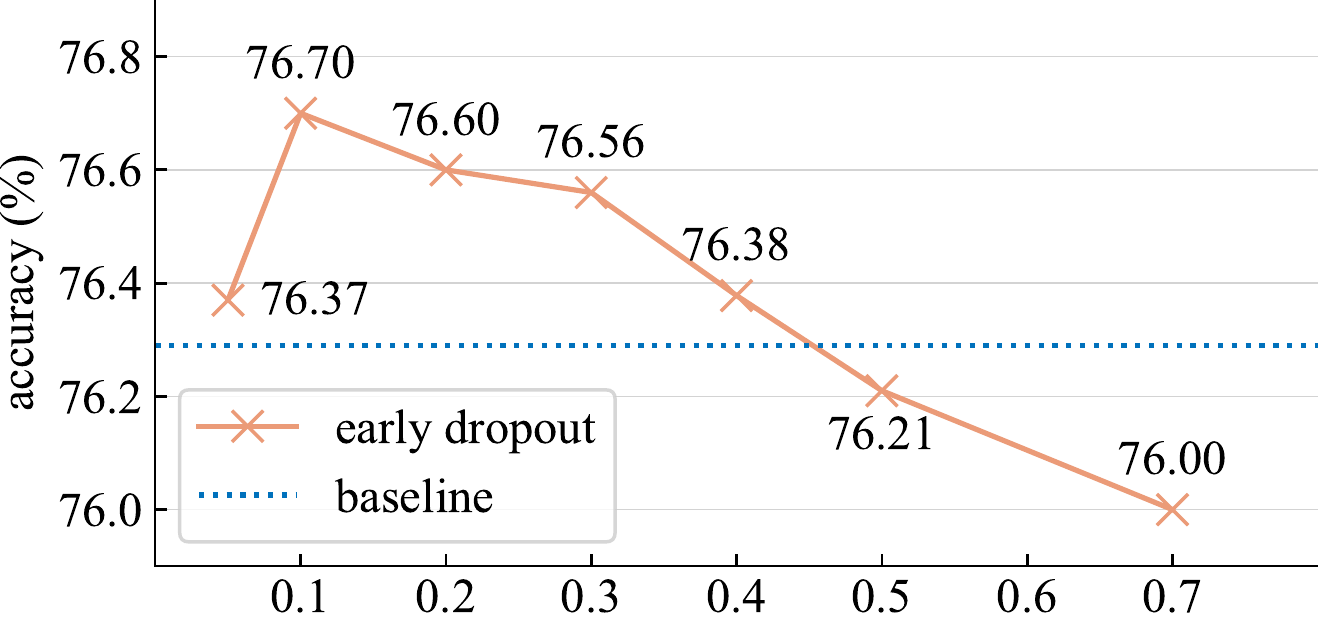}\\
\scriptsize early dropout rate \\
\includegraphics[width=.98\linewidth]{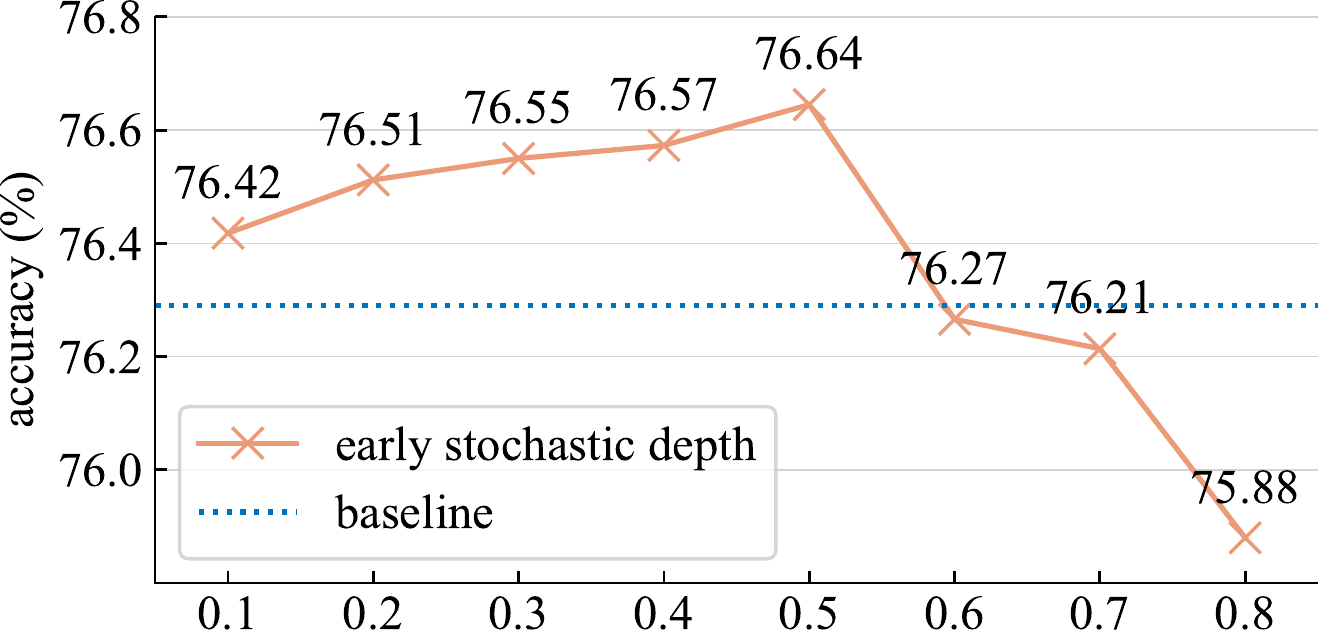}\\
\scriptsize early stochastic depth rate \\
\vspace{-.4em}
\caption{\textbf{Drop rates}. The performance of early dropout on \vit{} is affected by the dropout rate (top) but is more stable with the stochastic depth rate (bottom).
}
\vspace{-2em}
\label{fig:abl_rate}
\end{figure}

% \vspace{1em}
\paragraph{Drop rates.}
The dropout rate is another hyper-parameter, similar to standard dropout. The impact of varying the rate for \edo{} and \esd{} is shown in Figure~\ref{fig:abl_rate}. 
 The results indicate that the performance of \esd{} is not that sensitive to the rate, but the performance of \edo{} is highly dependent on it. This could be related to the fact that dropout layers are more densely inserted in ViTs than s.d. layers. In addition, the s.d. rate represents the maximum rate among layers~\cite{Huang2016deep}, but the dropout rate represents the same rates for all layers, so the same increase in dropout rate results in a much stronger regularizing effect. Despite that, both \edo{} and \esd{} are less sensitive to the rate than standard dropout, where a drop rate of 0.1 can significantly degrade accuracy  (Table~\ref{tab:main-early}).

\begin{table*}[t]
% \vspace{-.2em}
\centering
%#################################################
% strategy
%#################################################
\subfloat[
\textbf{Scheduling strategies}. Early dropout outperforms alternative strategies.
\label{tab:abl_strategy}
]{
\centering
\begin{minipage}{0.29\linewidth}{\begin{center}
\tablestyle{4pt}{1.05}
\begin{tabular}{x{30}x{30}x{33}}
strategy & acc. & train loss \\
\shline
\text{no dropout} & 76.3 & 3.033 \\
constant & 71.5 & 3.437 \\
\hline \\ [-2.3ex]
increasing & 75.2 & 3.285 \\
decreasing & 74.7 & 3.113 \\
annealed & 76.3 & 3.004 \\
curriculum & 70.4 & 3.490 \\
early & \textbf{76.7} & 2.996 \\
\end{tabular}
\end{center}}\end{minipage}
}
\hspace{2em}
%#################################################
% schedule
%#################################################
\subfloat[
\textbf{Early dropout scheduling}. Early dropout is robust to various schedules.
\label{tab:abl_schedule}
]{
% \centering
\begin{minipage}{0.29\linewidth}{\begin{center}
\tablestyle{4pt}{1.05}
\begin{tabular}{x{30}x{30}x{33}}
schedule & acc. & train loss \\
\shline \\ [-2.3ex]
linear & \textbf{76.7} & 2.991 \\
constant & 76.6 & 3.025 \\
cosine & 76.6 & 2.988 \\
 & & \\
\multicolumn{3}{c}{~}\\
\end{tabular}
\end{center}}\end{minipage}
}
\hspace{2em}
%#################################################
% model size
%#################################################
\subfloat[
\textbf{Model size}. Early dropout does not help models at overfitting regimes. 
% The baseline \vitb{} uses standard dropout.
\label{tab:abl_size}
]{
\begin{minipage}{0.29\linewidth}{\begin{center}
\tablestyle{1pt}{1.05}
\begin{tabular}{y{30}x{30}x{50}}
model & baseline & \edo{} \\
\shline \\ [-2.3ex]
\vit{} & 76.3 & \textbf{76.7} \\
ViT-S & 80.4 & \textbf{80.8} \\
\vitb{} & 78.7 & 78.7 \\
 & & \\
\multicolumn{3}{c}{~}\\
\end{tabular}
\end{center}}\end{minipage}
}
\vspace{-.3em}
\caption{\textbf{Early dropout ablation results} with \vit{}/16 on ImageNet-1K.}
\label{tab:ablations} 
\vspace{-.3em}
\end{table*}

\paragraph{Scheduling strategies.}
In previous studies, different strategies for scheduling dropout or related regularizers have been explored. These strategies typically involve either gradually increasing~\cite{morerio2017curriculum, Zoph2018, tan2021efficientnetv2} or decreasing~\cite{rennie2014annealed} the strength of dropout over the entire or nearly the entire training process. The purpose of these strategies, however, is to reduce overfitting rather than underfitting. 

For comparison, we also evaluate linear decreasing / increasing strategies where the drop rate starts from $p$ / 0 and ends at 0 / $p$, as well as previously proposed curriculum~\cite{morerio2017curriculum} and annealed~\cite{rennie2014annealed} strategies. For all strategies, we conduct a hyper-parameter sweep for the rate $p$. The results are presented in Table~\ref{tab:abl_strategy}. All strategies produce either similar or much worse results than no-dropout. This suggests existing dropout scheduling strategies are not effective for underfitting.

\paragraph{Early dropout scheduling.} 
There is still a question on how to schedule the drop rate in the early phase. Our experiments use a linear decreasing schedule from an initial value $p$ to 0 by default. A simpler alternative is to use a constant value. It can also be useful to consider a cosine decreasing schedule commonly adopted for learning rate schedules. The optimal $p$ value for each option may differ and we compare the best result for each option. Table~\ref{tab:abl_schedule} presents the results. All three options manifest similar results and can serve as valid choices. This indicates early dropout does not depend on one particular schedule to work. Additional results for constant early dropout can be found in Appendix~\ref{appendix:con}.

\paragraph{Model sizes.}
According to our analysis in Section~\ref{sec:how}, early dropout helps models fit better to the training data. This is particularly useful for underfitting models like \vit{}. We take ViTs of increasing sizes, \vit{}, ViT-S, and \vitb{}, and examine the trend in Table~\ref{tab:abl_size}. The baseline column represents the results obtained by the best standard dropout rates (0.0 / 0.0 / 0.1) for each of the three models. 
Our results show that \edo{} is effective in improving the performance of the first two models, but was not effective in the case of the larger \vitb{}.

\begin{figure}[h]\centering
\includegraphics[width=.98\linewidth]{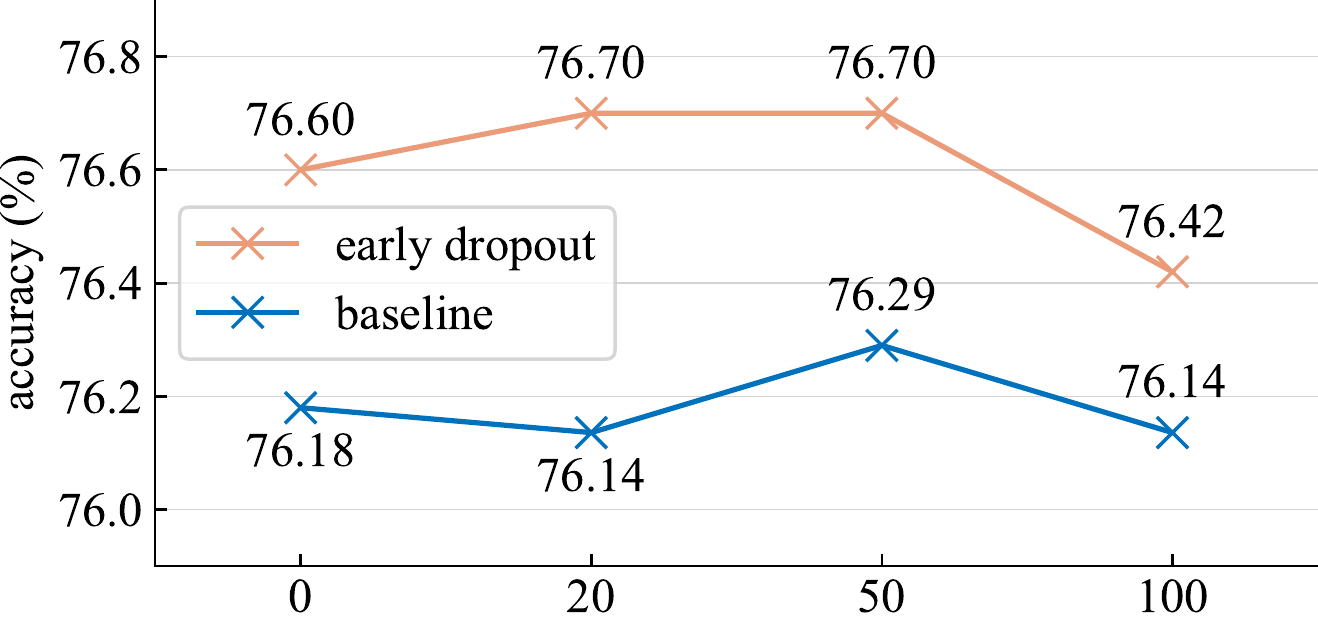}\\
\scriptsize learning rate warmup epochs\\
\vspace{-.5em}
\caption{Early dropout leads to accuracy improvement when the number of \textbf{learning rate warmup} epochs varies.}
\label{fig:abl_lrwarmup}
\vspace{-1em}
\end{figure}

\paragraph{Learning rate warmup.}
Learning rate (lr) warmup~\cite{He2016,Goyal2017} is a technique that also specifically targets the early phase of training, where a smaller lr is used. We are curious in the effect lr warmup on early dropout. Our default recipe uses a 50-epoch linear lr warmup. We vary the lr warmup length from 0 to 100 and compare the accuracy with and without early dropout in Figure~\ref{fig:abl_lrwarmup}. Our results show that early dropout consistently improves the accuracy regardless of the use of lr warmup.

\begin{figure}[h]\centering
\includegraphics[width=.98\linewidth]{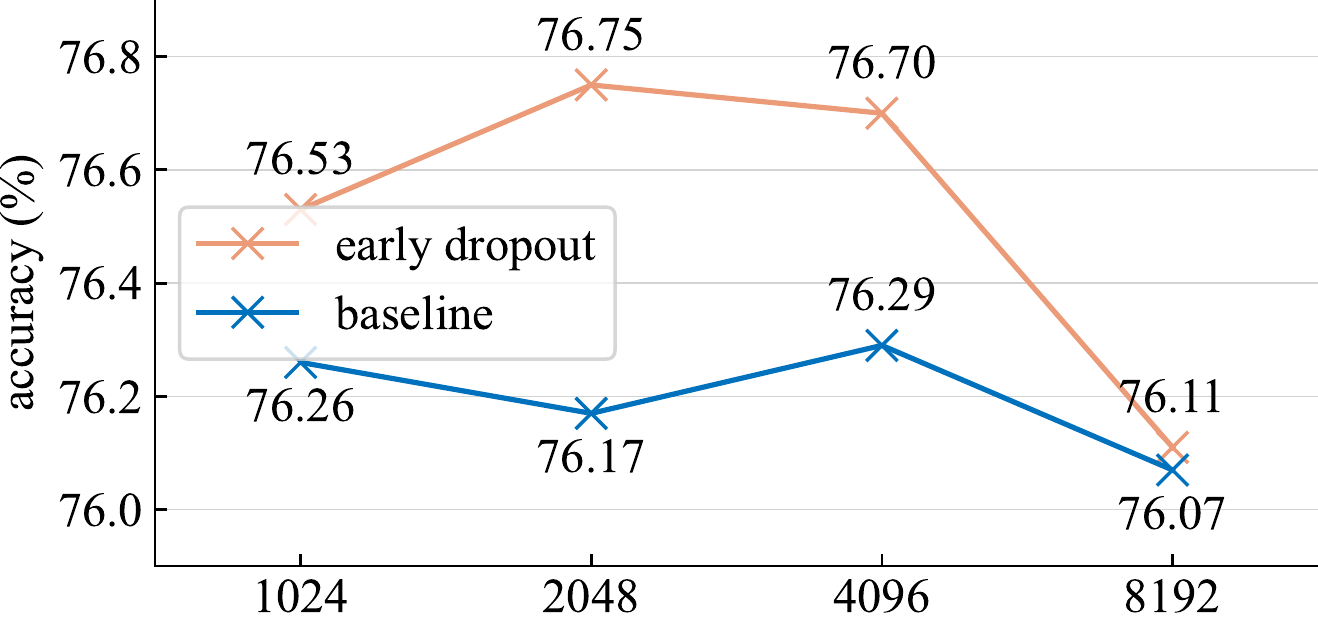}\\
\scriptsize Batch size\\
\vspace{-.5em}
\caption{Early dropout is not as effective when the batch size is increased to 8192, but consistent improvement is observed for smaller batch sizes. This supports our hypothesis on the gradient error reduction effect of \edo{}.}
\label{fig:abl_bs}
% \vspace{-1em}
\end{figure}

\paragraph{Batch size.}
We vary the batch size from 1024 to 8192 and scale the learning rate linearly \cite{Goyal2017} to examine how batch size influences the effect of \edo{}. Our default batch size is set at 4096. In Figure~\ref{fig:abl_bs}, we note that early dropout becomes less beneficial as the batch size increases to 8192. This observation supports our hypothesis: as the batch size grows, the mini-batch gradient tends to approximate the entire-dataset gradient more closely. Consequently, the importance of gradient error reduction may diminish, and \edo{} no longer yields meaningful improvement over the baseline.

\begin{figure}[h]\centering
\includegraphics[width=0.99\linewidth]{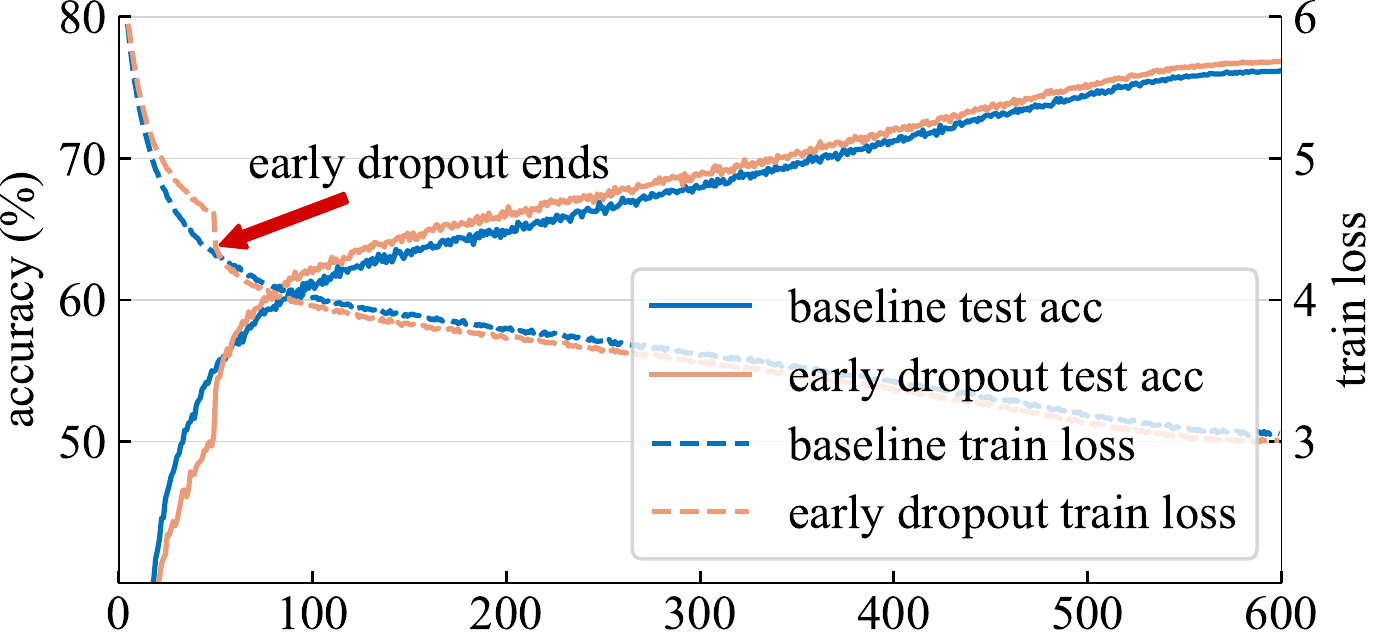}\\
\scriptsize \,\,\,\, epochs \\
\vspace{-.7em}
\caption{\textbf{Training Curves.} 
When early dropout ends, the model experiences a significant decrease in training loss and a corresponding increase in test accuracy.
}
\label{fig:curve}
\vspace{-.9em}
\end{figure}

\paragraph{Training curves.} We plot the training loss and test accuracy curves for \vit{} with \edo{} and compare it with a no-dropout baseline in Figure~\ref{fig:curve}. The early dropout is set to 50 epochs and uses a constant dropout rate. During the \edo{} phase, the train loss for the dropout model is higher and the test accuracy is lower. Intriguingly, once the early dropout phase ends, the train loss decreases dramatically and the test accuracy improves to surpass the baseline.

\subsection{Late Dropout}
\paragraph{Settings.} 
To evaluate \ldo{}, we choose larger models, \vitb{} and \mixerb{}, with 59M and 86M parameters respectively, and use the basic training recipe. These models are considered to be in the overfitting regime as they already use standard s.d.
We evaluate late s.d. because we find the baseline results using standard s.d. are much better than standard dropout for these models.
For this experiment, we set the drop rate for \lsd{} directly to their optimal drop rate for \csd{} 
No s.d. is used for the first 50 epochs, and a constant s.d. rate is used for the rest of training.

\paragraph{Results.} 
In the results shown in Table~\ref{tab:main-late}, late s.d. improves the test accuracy compared to \csd{}.  
This improvement is achieved while either maintaining (\vitb{}) or increasing (\mixerb{}) the training loss, demonstrating that \lsd{} effectively reduces overfitting. Previous works~\cite{morerio2017curriculum,tan2021efficientnetv2,Zoph2018} have used dropout with gradually increasing strength to combat overfitting. In the case of \vitb{}, we also compare our results with a linear increase and a curriculum schedule~\cite{morerio2017curriculum} with their best $p$ over a hyperparameter sweep and find that \lsd{} brings a larger improvement. Appendix~\ref{appendix:late} presents more detailed analysis for late s.d.

\begin{table}[t]
\centering
\small
\addtolength{\tabcolsep}{-5.pt}
\def\arraystretch{1.1}% 
\vspace{-.2em}
\begin{tabular}{lcccc}
   model &  top-1 acc. & change  & train loss & change  \\
\Xhline{1.0pt}
\vitb{} (\csd{})$^*$  & 81.8  & -            & - & -                 \\
\vitb{} (\csd{})     & 81.6  & -            & 2.817 & -                 \\
+ \nsd{}             & 77.0  & \worse{4.8}            & 2.255 & \betterinv{0.562}                 \\
+ linear-increasing s.d.            & 82.1  & \better{0.5}            & 2.939 & \worseinv{0.122}                 \\
+ curriculum$^\ddagger$ s.d.            & 82.0  & \better{0.4}            & 2.905 & \worseinv{0.088}                 \\
\gr + \lsd{}         & \textbf{82.3}  & \better{0.7}            & 2.808 & \betterinv{0.009}                 \\

\hline
\mixerb{} (\csd{})$^\dagger$ & 76.4  & -            & - & -                 \\
\mixerb{} (\csd{})    & 78.0  & -            & 2.810 & -                 \\
+ \nsd{}             & 76.0  & \worse{2.0}           & 2.468 & \betterinv{0.342}                 \\
\gr + \lsd{}         & \textbf{78.6}  & \better{0.6}            & 2.865 & \worseinv{0.055}                 \\

\hline
\end{tabular}
\normalsize
% \vspace{-.6em}
\caption{\textbf{Classification accuracy on ImageNet-1K for late s.d.}  Late s.d. leads to improved test accuracy for overfitting models compared to their standard counterparts. Literature baselines: $*$\citet{Touvron2020}, $\dagger$\citet{tolstikhin2021mlp}.}
\vspace{-.5em}
\label{tab:main-late}
\end{table}

\section{Downstream Tasks}
\label{sec:downstream}
We evaluate the pre-trained ImageNet-1K models by fine-tuning them on downstream tasks. Our aim is to evaluate the learned representations without using early or late dropout during fine-tuning. Additionally, we conduct a direct evaluation of robustness benchmarks in Appendix~\ref{appendix:rob}.

\paragraph{Object detection and segmentation on COCO.} We fine-tune pre-trained \swin{} and \convnext{} backbones with Mask-RCNN~\cite{He2017} on the COCO dataset. We use the 1$\times$ fine-tuning setting in MMDetection~\cite{mmdetection}.
We follow the 1$\times$ fine-tuning setting in MMDetection~\cite{mmdetection}. The results are shown in Table~\ref{tab:downstream_det}. Models pre-trained with early dropout or s.d.  consistently maintain their superiority when fine-tuned on COCO.

\begin{table}[h]
\tablestyle{6pt}{1.1}
\addtolength{\tabcolsep}{-2.5pt}
\vspace{2ex}
\scalebox{0.97}{
\begin{tabular}{@{}lcccccc@{}}
backbone & $\text{AP}^{\text{box}}$ & $\text{AP}^{\text{box}}_{50}$ & $\text{AP}^{\text{box}}_{75}$ & $\text{AP}^{\text{mask}}$ & $\text{AP}^{\text{mask}}_{\text{50}}$ & $\text{AP}^{\text{mask}}_{75}$  \\
\shline
\multicolumn{7}{c}{\scriptsize{Mask-RCNN 1$\times$ schedule}} \\

\swin{}            & 36.4 & 58.8 & 38.8 & 34.2 & 55.6 & 36.0 \\
\gr + \edo{}      & \textbf{37.1} & 59.1 & 39.6 & \textbf{34.6} & 56.0 & 36.5 \\
\gr + \esd{}      & \textbf{36.9} & 59.3 & 39.4 & \textbf{34.5} & 56.1 & 36.4 \\

\Xhline{0.3\arrayrulewidth}
\convnext{}            & 46.0 & 68.1 & 50.3 & 41.6 & 65.1 & 44.9 \\
\gr + \esd{}      & \textbf{46.2} & 67.9 & 50.8 & \textbf{41.7} & 65.0 & 44.9 \\

\end{tabular}
} 
% \vspace{-.4em}
\caption{\textbf{COCO object detection and segmentation results}.
\label{tab:downstream_det}
}
\end{table}

\paragraph{Semantic segmentation on ADE20K.} We fine-tune pre-trained models on the ADE-20K semantic segmentation task ~\cite{Zhou2019} with UperNet~\cite{Xiao2018} for 80k iterations, following MMSegmentation~\cite{mmseg2020}. As Table~\ref{tab:ade20k} shows, models pre-trained with our methods outperform baseline models.
%##################################################################################################
\begin{table}[t]
% \vspace{-.5em}
\tablestyle{8pt}{1.12}
\begin{tabular}{lcc}
method & \vit{} & \vitb{} \\
\shline
baseline &  39.2 & 44.3 \\
\gr + \edo{} &  \textbf{40.0} & - \\
\gr + \esd{} &  \textbf{39.8} & - \\
\gr + \lsd{} &  - & \textbf{45.7} \\
\end{tabular}
% \vspace{-.4em}
\caption{\textbf{ADE20K semantic segmentation results (mIoU).}}
\label{tab:ade20k} 
\vspace{-.5em}
\end{table}
%##################################################################################################

\paragraph{Downstream classification tasks.}
We also evaluate model fine-tuning on several downstream classification datasets: CIFAR-100~\cite{Krizhevsky2009}, Flowers~\cite{Nilsback2008}, Pets~\cite{Parkhi2012}, STL-10~\cite{coates2011analysis} and Food-101~\cite{bossard2014food}. Our fine-tuning procedures are based on the hyper-parameter settings from MoCo v3~\cite{Chen2021a} and SLIP~\cite{mu2022slip}. Table~\ref{tab:downstream_cls} presents the results. Our methods show improved  performance on most classification tasks.

\begin{table}[h]
\tablestyle{8.8pt}{1.1}
\addtolength{\tabcolsep}{-1.5pt}
\scalebox{0.9}{
\begin{tabular}{lccccc}
Model &  C-100 & Flowers & Pets & STL-10 & F-101 \\
\toprule
\vit{}  & 87.4 & 96.2 & 92.2 & 97.6 & 89.7 \\

\gr + \edo{} & \textbf{87.9} & \textbf{96.4} & \textbf{93.1} & \textbf{97.8} & \textbf{89.9}  \\

\Xhline{0.3\arrayrulewidth}

\swin{}  & 86.5 & 96.2 & 92.2 & 97.7 & 89.4  \\

\gr + \edo{} & \textbf{86.9} & \textbf{96.7} & \textbf{92.3} & \textbf{97.8} & \textbf{89.5}  \\

\Xhline{0.3\arrayrulewidth}
\vitb{}$^*$ & 87.1 & 89.5 & 93.8 & - & - \\
\vitb{}$^\dagger$ & 90.5 & 97.7 & 93.2 & - & - \\
\vitb{}  & 90.5 & 97.5 & \textbf{95.4} & 98.5 & 90.6 \\
\gr + \lsd{} & \textbf{90.7} & \textbf{97.9} & 95.3 & \textbf{98.7} & \textbf{91.4}  \\

\end{tabular}
} 
\caption{\textbf{Downstream classification accuracy} on five datasets. Literature baselines: $*$\citet{Dosovitskiy2021}, $\dagger$\citet{Chen2021a}. }
\vspace{-.5em}
\label{tab:downstream_cls}
\end{table}

\section{Related Work}
\label{sec:related}

%%%%%%%%%%%%
\paragraph{Neural network regularizers.} 
Weight decay, or $L_2$ regularization, is one of the most commonly used regularization for training neural networks. Related to our findings, \citet{Krizhevsky2012} observe that using weight decay decreases the training loss for AlexNet. $L_1$ regularization~\cite{tibshirani1996regression} can promote sparsity and select features~\cite{liu2017learning}.
Label smoothing~\cite{Szegedy2016a} replaces one-hot targets output with soft probabilities. 
Data augmentation~\cite{Zhang2018a,Cubuk2020} can also serve as a form of regularization. In particular, methods that randomly remove input parts, e.g., hide-and-seek~\cite{kumar2017hide}, cutout~\cite{devries2017improved} and random ereasing~\cite{Zhong2020}, can be seen as dropout applied at the input layer only. 
%%%%%%%%%

\paragraph{Dropout methods.}
Dropout has many variants aimed at improving or adapting it. DropConnect~\cite{Wan2013} randomly deactivates network weights instead of neurons. Variational dropout~\cite{kingma2015variational} adaptively learns dropout rates for different parts of the network from a Bayesian perspective. Spatial dropout~\cite{tompson2015efficient} drops entire feature maps in a ConvNet, and DropBlock~\cite{ghiasi2018dropblock} drops continuous regions in ConvNet feature maps. 
Other valuable contributions include analyzing dropout properties~\cite{baldi2013understanding,ba2013adaptive,wang2013fast}, applying dropout for compressing networks~\cite{molchanov2017variational,gomez2019learning} and representing uncertainty~\cite{gal2016dropout,gal2017concrete}. 
We recommend the survey by \citet{labach2019survey} for a comprehensive overview.

\paragraph{Scheduled dropout.}
Neural networks generally tend to show overfitting behaviors more at later stages of training, which is why \emph{early stopping} is often used to reduce overfitting. 
Curriculum dropout~\cite{morerio2017curriculum} proposes to increase the dropout rate as training progresses to more specifically address late-stage overfitting.
NASNet~\cite{Zoph2018} and EfficientNet-V2~\cite{tan2021efficientnetv2} also increase the strength of dropout / drop-path~\cite{larsson2016fractalnet} during neural architecture search.
On the other hand, annealed dropout~\cite{rennie2014annealed} gradually decreases dropout rates to near the end of training. 
Our approaches differ from previous research as we study dropout's effect in addressing \emph{underfitting} rather than regularizing overfitting. 

\section{Conclusion}
\label{sec:conclusion}
Dropout has shined for 10 years for its excellence in tackling overfitting. In this work, we unveil its potential in aiding stochastic optimization and reducing underfitting. Our key insight is dropout counters the data randomness brought by SGD and reduces gradient variance at early training. 
This also results in stochastic mini-batch gradients that are more aligned with the underlying whole-dataset gradient. Motivated by this, we propose \edo{} to help underfitting models fit better, and \ldo{}, to improve the generalization of overfitting models. 
We hope our discovery stimulates more research in understanding dropout and designing regularizers for gradient-based learning, and our approaches help model training with increasingly large datasets.

\paragraph{Acknowledgement.}
We would like to thank Yubei Chen, Yida Yin, Hexiang Hu, Zhiyuan Li, Saining Xie and Ishan Misra for valuable discussions and feedback.

\bibliography{main}
\bibliographystyle{icml2023}
%%%%%%%%%%%%%%%%%%%%%%%%%%%%%%%%%%%%%%%%%%%%%%%%%%%%%%%%%%%%%%%%%%%%%%%%%%%%%%%
%%%%%%%%%%%%%%%%%%%%%%%%%%%%%%%%%%%%%%%%%%%%%%%%%%%%%%%%%%%%%%%%%%%%%%%%%%%%%%%
% APPENDIX
%%%%%%%%%%%%%%%%%%%%%%%%%%%%%%%%%%%%%%%%%%%%%%%%%%%%%%%%%%%%%%%%%%%%%%%%%%%%%%%
%%%%%%%%%%%%%%%%%%%%%%%%%%%%%%%%%%%%%%%%%%%%%%%%%%%%%%%%%%%%%%%%%%%%%%%%%%%%%%%
\clearpage
\appendix

% \vspace*{0.0em}
\section*{\Large{Appendix}}
\vspace{0.5em}
\section{Experimental Settings}
\label{appendix:exp}

\textbf{Training recipe.} We provide our basic training recipe with specific details  in Table~\ref{tab:train_detail}. This recipe is based on the setting in ConvNeXt~\cite{liu2022convnet}. For the improved recipe, we increase the number of epochs to 600, and reduce mixup and cutmix to 0.3. All other configurations remain unchanged.

\textbf{Drop rates.} The drop rates for \edo{} and \esd{} are listed in Table \ref{tab:hp_early}. By default, the early dropout epochs are set to 50, with a linear decreasing schedule. A light search of early dropout rates was conducted from the values \{0.1, 0.2, 0.3\}. For \swin{}, we find including an additional range \{0.5, 0.7\} is useful. For early s.d. rate, we search from \{0.3, 0.5, 0.7\} for all models. The baselines do not use any dropout or s.d. The compared standard dropout / s.d. experiments all use a low drop rate of 0.1.

The late s.d. drop rates are listed in Table~\ref{tab:hp_late}. The basic training recipe is adopted. The baselines use standard s.d., whose rates are obtained with hyper-parameter sweeps. We find using the same rates for late s.d. proves to be effective.

\begin{table*}[b]

\tablestyle{7.0pt}{1.07}
\footnotesize
% \vspace{-3em}
\begin{tabular}{@{\hskip 2ex}l|c@{\hskip 2ex}c} \\
% \vspace{-8em}
Training Setting & Configuration \\
\shline
weight init & trunc. normal (0.2)  \\
optimizer & AdamW \\
base learning rate & 4e-3  \\
weight decay & 0.05  \\
optimizer momentum & $\beta_1, \beta_2{=}0.9, 0.999$  \\
batch size & 4096  \\
training epochs & 300  \\
learning rate schedule & cosine decay  \\
warmup epochs & 50  \\
warmup schedule & linear  \\
\textbf{stochastic depth rate} \cite{Huang2016deep} & 0.0 \\
\textbf{dropout rate} \cite{Hinton2012} & 0.0 \\
randaugment \cite{Cubuk2020} & (9, 0.5)  \\
mixup \cite{Zhang2018a} & 0.8  \\
cutmix \cite{Yun2019} & 1.0  \\
random erasing \cite{Zhong2020} & 0.25  \\
label smoothing \cite{Szegedy2016a} & 0.1  \\
layer scale \cite{Touvron2021GoingDW} & 1e-6  \\
gradient clip & None  \\
exp. mov. avg. (EMA) \cite{Polyak1992} & None \\
\end{tabular}
\caption{Our basic training recipe, adapted from ConvNeXt~\citep{liu2022convnet}.}
\vspace{6em}
\label{tab:train_detail}
\end{table*}

% \newpage
\begin{table}[h]
\tablestyle{8pt}{1.0}
\vspace{.5em}
\begin{tabular}{lcc}
model & early dropout rate & early s.d. rate \\
\shline
\multicolumn{3}{c}{\scriptsize{with basic recipe}} \\
\vit{} &  0.1 & 0.5 \\
\mixer{} &  0.1 & 0.7 \\
\convnext{} &  - & 0.5 \\
\swin{} &  0.5 & 0.5 \\
\hline
\multicolumn{3}{c}{\scriptsize{with improved recipe}} \\
\vit{} &  0.1 & 0.7 \\
\convnext{} &  - & 0.5 \\
\swin{} &  0.7 & 0.5 \\
\end{tabular}
\caption{Early dropout and early s.d. rates used in experiments.}
\label{tab:hp_early} 
\end{table}

\begin{table}[h]
\tablestyle{8pt}{1.0}
\vspace{-8em}
\begin{tabular}{lcc}
model & standard s.d. rate & early s.d. rate \\
\shline
\multicolumn{3}{c}{\scriptsize{with basic recipe}} \\
\vitb{} &  0.4 & 0.4 \\
\mixerb{} &  0.2 & 0.2 \\
\end{tabular}
\caption{Late s.d. rates and standard s.d. rates used in experiments.}
\label{tab:hp_late} 
\end{table}

% \newp
\clearpage
\clearpage

\section{Analaysis for Late Dropout}
\label{appendix:late}

\begin{figure}[h]\centering
\includegraphics[width=0.99\linewidth]{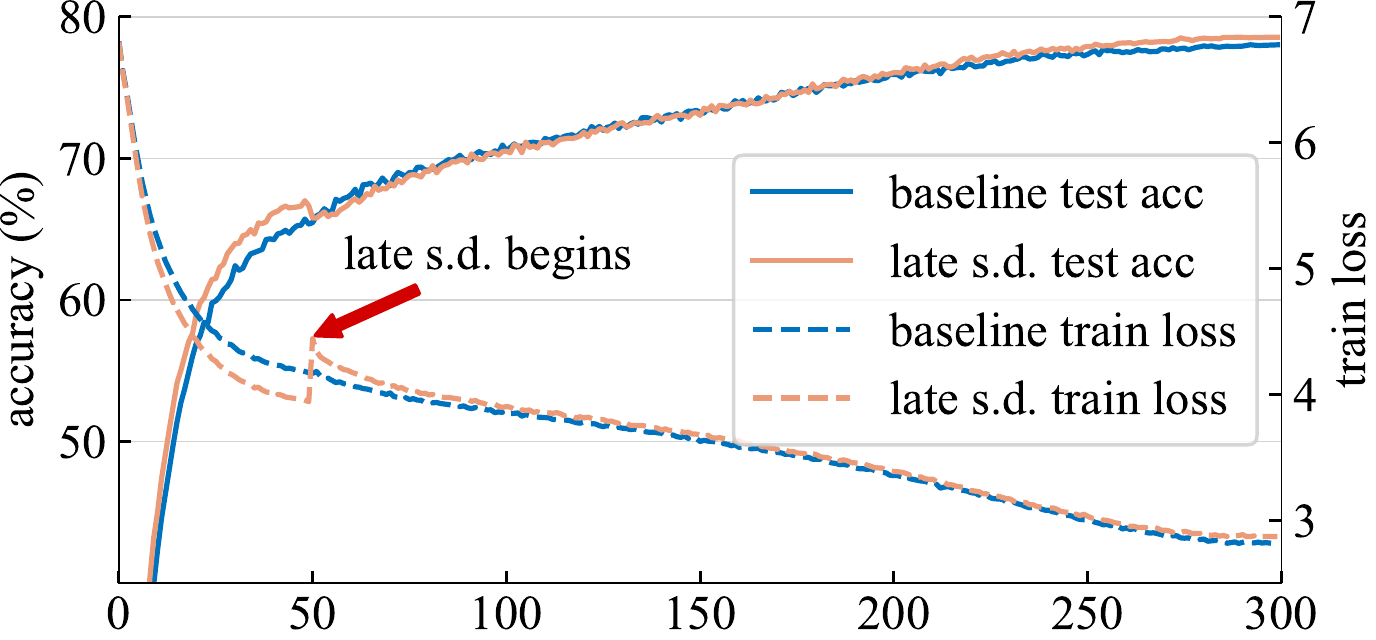}\\
% \vspace{-.5em}
\scriptsize{} epochs
\vspace{-.5em}
\caption{\textbf{Training Curves.}
\label{fig:late_curve}
When late s.d. begins, the model experiences a jump in training loss and a decrease in test accuracy.
}
\end{figure}

\paragraph{Training curves.} We present the training curves for late s.d. in Figure~\ref{fig:late_curve}, comparing it with the baseline (standard s.d. with the best drop rate). When late s.d. begins, the training loss immediately increases. However, the final test accuracy of the late s.d. model is higher than the baseline and so is the training loss, demonstrating the effectiveness of late s.d. in reducing overfitting and closing the generalization gap.

\begin{figure}[h]\centering
\vspace{-.5em}
\includegraphics[width=.98\linewidth]{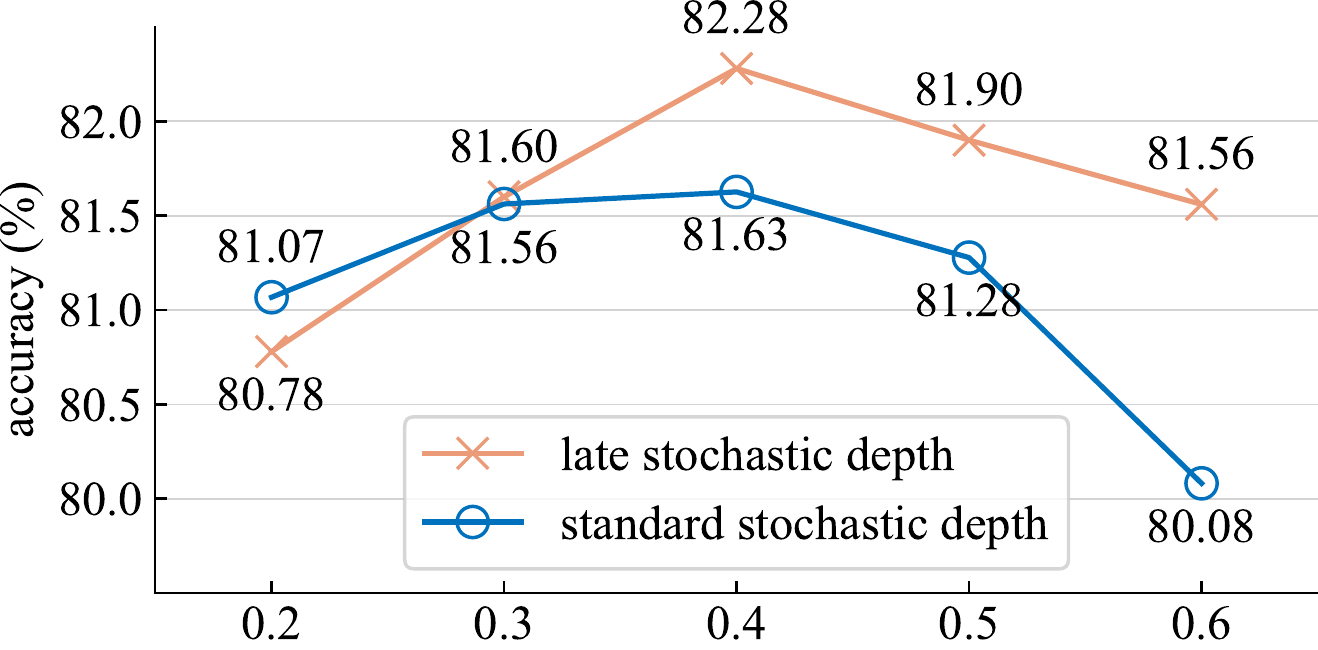}\\
\scriptsize late stochastic depth rate \\
\vspace{-.5em}
\caption{\textbf{Late s.d. drop rates}. Late s.d. improves over standard s.d. for a broad range of drop rates.}
\label{fig:late_abl_rate}
\end{figure}

\paragraph{Drop rates.} We examine the impact of the drop rate for late s.d. As the models are in an overfitting regime, we also plot the results using different standard s.d. rates as baselines. In Figure~\ref{fig:late_abl_rate}, we observe that late s.d. is less sensitive to changes in the drop rate and, overall, leads to improved generalization results. The only s.d. rate where \lsd{} hurts the performance is 0.2, which is suboptimal for the baseline too.

\begin{figure}[h]\centering
\hspace{-1em}
\includegraphics[width=.98\linewidth]{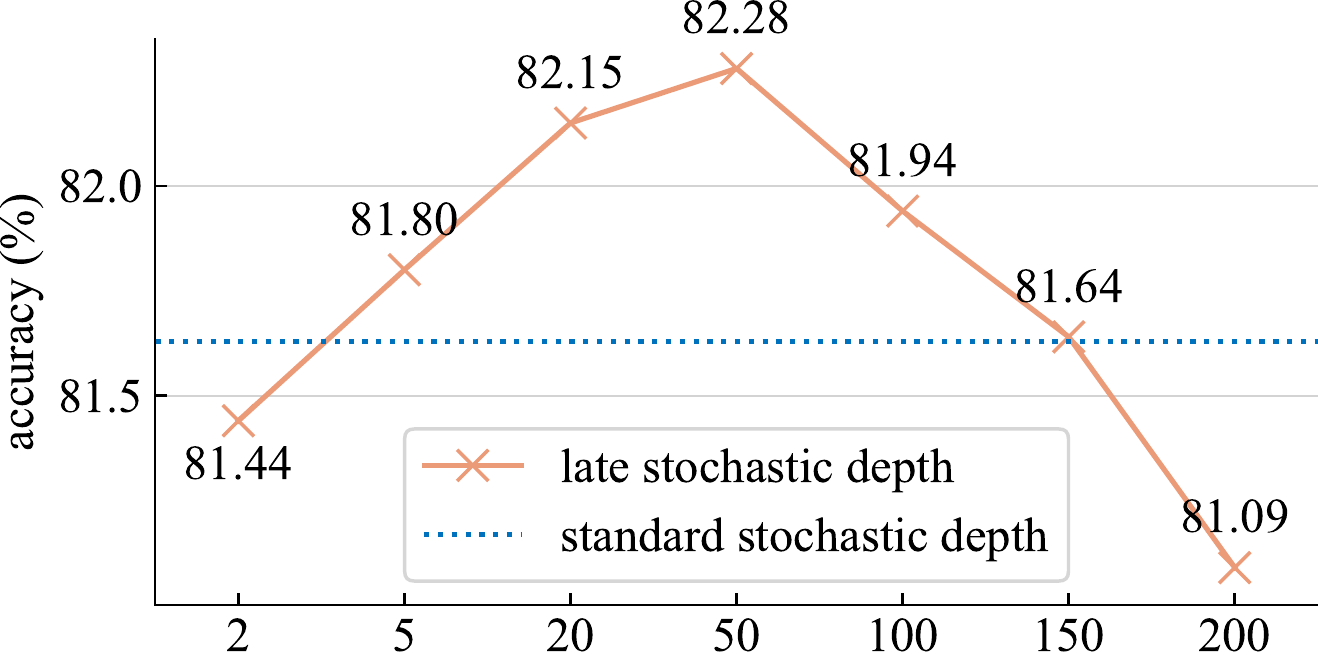}\\
\scriptsize late stochastic depth epochs \\
\caption{\textbf{Late s.d. epochs}. The optimal epoch for late s.d. in this experiment is 50.
}
\label{fig:late_abl_epoch}
\end{figure}

% \newpage
\paragraph{Dropout epochs.} Similarly, we analyze the effect of different late s.d. epochs in Figure~\ref{fig:late_abl_epoch}. The epoch refers to the point where s.d. begins. Overall, the improvement from late s.d. remains consistent when the start epoch varies from 5 to 100, with a peak observed at 50. The optimal epoch for late s.d. may vary based on the chosen drop rate.

\paragraph{Other architectures.} We attempted to use late s.d. on ConvNeXt-B and Swin-B, but were unable to find a set of hyper-parameters that resulted in a significant improvement over standard s.d.The differing results compared to those obtained with \vitb{} and \mixerb{} could be attributed to the inductive biases present in these architectures. Further investigation is needed to determine why late s.d. may not be suitable for certain architectures.

\section{Standard Deviation Results}
We provide standard deviation details corresponding to Table~\ref{tab:main-early} below. Each experiment employs 3 random seeds. The improvement in mean accuracy generally exceeds the standard deviation, indicating reliable \edo ~enhancements across models, dropout variants, and training recipes.

\begin{table}[h]
\centering
\small
\addtolength{\tabcolsep}{7.pt}
\def\arraystretch{1.2}% 
\begin{tabular}{lc}
   model & top-1 acc.  \\
\Xhline{1.0pt}
\multicolumn{2}{c}{\scriptsize{results with basic recipe}}   \\
\vit{}      & 73.89 $\pm$ 0.20               \\
\gr+ \edo{} & \bb{74.26} $\pm$ 0.13  \\
\gr+ \esd{} & \bb{74.38} $\pm$ 0.14 \\
\hline
\mixer{}    & 70.95  $\pm$    0.15             \\
\gr+ \edo{} & \bb{71.29} $\pm$ 0.22 \\
\gr+ \esd{} & \bb{71.74} $\pm$ 0.24 \\

\hline
\convnext{} & 76.11   $\pm$   0.22        \\
\gr+ \esd{} & \bb{76.33} $\pm$  0.03 \\

\hline
\swin{}     & 74.27 $\pm$  0.08  \\
\gr+ \edo{} & \bb{74.68} $\pm$ 0.18 \\
\gr+ \esd{} & \bb{75.15} $\pm$  0.07 \\
\hline

\multicolumn{2}{c}{\scriptsize{results with improved recipe}}  \\ 
\vit{}      & 76.29  $\pm$  0.17 \\
\gr+ \edo{} & \bb{76.70}  $\pm$  0.02 \\
\gr+ \esd{} & \bb{76.67} $\pm$   0.17 \\

\hline
\convnext{} & 77.48 $\pm$  0.12 \\
\gr+ \esd{} & \bb{77.67} $\pm$ 0.13 \\

\hline
\swin{}     & 76.07  $\pm$ 0.13 \\
\gr+ \edo{} & \bb{76.55}  $\pm$ 0.20 \\
\gr+ \esd{} & \bb{76.63} $\pm$  0.11 \\
\hline
\end{tabular}
\normalsize
\caption{\textbf{Main results with standard deviation.}}
\label{tab:main-add}
\end{table}

\clearpage
% \newpage
\section{Constant Early Dropout}
\label{appendix:con}
The majority of experiments described in paper use a linear decreasing schedule for early dropout. We now switch to a constant schedule, where the early dropout phase uses a constant drop rate, and then turned off to 0 when it ends. This is also discussed in Table~\ref{tab:abl_schedule}'s experiments.

We find it beneficial to shorten the dropout epochs from 50 to 20. This is perhaps because the ``accumulated'' drop rate (calculated as the area under the curve on a drop rate vs. epoch plot) plays an important role, and constant schedule accumulates twice as much as the linear schedule if they both start at the same rate $p$ and end at the same epoch.

We present the results in Table~\ref{tab:main-add}. Constant early dropout consistently improves both training loss and test accuracy upon the baseline. This further demonstrates that early dropout is not limited to a linearly decreasing schedule to effectively reduce underfitting.

\begin{table}[h]
\centering
\small
\addtolength{\tabcolsep}{-3.pt}
\def\arraystretch{1.1}% 
% \vspace{-1em}
\begin{tabular}{lcccc}
   model & top-1 acc. & change  &  train loss & change  \\
\Xhline{1.0pt}
\multicolumn{5}{c}{\scriptsize{results with basic recipe}}   \\
\vit{}      & 73.9  & -            & 3.443 & -                 \\
\gr+ \edo{} & \bb{74.4}  & \better{0.5} & 3.408 & \betterinv{0.035} \\
\gr+ \esd{} & \bb{74.0}  & \better{0.1} & 3.428 & \betterinv{0.015} \\

\hline
\mixer{}$^*$& 68.7  & -            & - & - \\
\mixer{}    & 71.0  & -            & 3.635 & -                 \\
\gr+ \edo{} & \bb{71.4}  & \better{0.4} & 3.572 & \betterinv{0.063} \\
\gr+ \esd{} & \bb{71.6}  & \better{0.6} & 3.553 & \betterinv{0.082} \\

\hline
\convnext{} & 76.1  & -            & 3.472 & -                 \\
\gr+ \esd{} & \bb{76.5}  & \better{0.4} & 3.449 & \betterinv{0.023} \\

\hline
\swin{}     & 74.3  & -            & 3.411 & -                 \\
\gr+ \edo{} & \bb{74.6}  & \better{0.3} & 3.382 & \betterinv{0.029} \\
\gr+ \esd{} & \bb{75.1}  & \better{0.8} & 3.355 & \betterinv{0.056} \\
\hline

\multicolumn{5}{c}{\scriptsize{results with improved recipe}}  \\  
\vit{}$^\dagger$ & 72.8  & -            & - & -                 \\
\vit{}$^\ddagger$ & 75.5  & -            & - & -                 \\
\vit{}      & 76.3  & -            & 3.033 & -                 \\
\gr+ \edo{} & \bb{76.7}  & \better{0.4} & 2.994 & \betterinv{0.043} \\
\gr+ \esd{} & \bb{76.7}  & \better{0.4} & 3.008 & \betterinv{0.025} \\

\hline
\convnext{}$^\ddagger$ & 77.5  & -            & - & -                 \\
\convnext{} & 77.5  & -            & 3.011 & -                 \\
\gr+ \esd{} & \bb{77.6}  & \better{0.1} & 2.989 & \betterinv{0.022} \\

\hline
\swin{}     & 76.1  & -            & 2.989 & -                 \\
\gr+ \edo{} & \bb{76.4}  & \better{0.3} & 2.972 & \betterinv{0.017} \\
\gr+ \esd{} & \bb{76.8}  & \better{0.7} & 2.974 & \betterinv{0.015} \\
\hline
\end{tabular}
\normalsize
\caption{\textbf{Classification accuracy on ImageNet-1K with early dropout using a constant schedule.} We obtain consistent improvement with results similar to those obtained using a linear schedule. Literature baselines: $*$\citet{tolstikhin2021mlp}, $\dagger$\citet{Touvron2020}, $\ddagger$\citet{rw2019timm}.
}
\label{tab:main-add}
\end{table}

\newpage
\section{Robustness Evaluation}
\label{appendix:rob}
We evaluate the models on common robustness benchmarks, which test their accuracy when the input images experience a change in distribution, such as corruption or style change. We report top-1 accuracy on ImageNet-A~\cite{hendrycks2021natural}, ImageNet-R~\cite{hendrycks2021many}, ImageNet-Sketch~\cite{wang2019learning}, ImageNet-V2~\cite{recht2019imagenet}, Stylized ImageNet~\cite{geirhos2018imagenet}, and mean Corruption Error (mCE) on ImageNet-C~\cite{hendrycks2018benchmarking}. Table~\ref{tab:robustness} shows that the improvement is transferable across different conditions.

\begin{table}[h]
\tablestyle{8.3pt}{1.15}
\addtolength{\tabcolsep}{-3.5pt}
\scalebox{0.9}{
\begin{tabular}{lccccccc}
Model &  Clean & A & R & SK & V2 & Style & C ($\downarrow$)\\
\shline

\vit{} & 76.3 & 10.2 & 36.3 & 24.2 & 63.7 & 12.3 & 65.4 \\

\gr + \edo{} & \bb{76.7} & \bb{11.6} & \bb{37.3} & \bb{24.7} & \bb{65.0} & \bb{13.0} & \bb{64.2} \\
\gr + \esd{} & \bb{76.7} & \bb{10.0} & \bb{36.8} & \bb{24.8} & \bb{64.2} & \bb{12.8} & \bb{63.6} \\

\Xhline{0.3\arrayrulewidth}

\mixer{} & 71.0 & 4.1 & 35.4 & 23.0 & 56.8 & 13.0 & 67.7 \\

\gr + \edo{} & \bb{71.3} & \bb{4.2} & \bb{35.9} & \bb{23.5} & \bb{58.2} & \bb{13.5} & \bb{66.3} \\
\gr + \esd{} & \bb{71.7} & \bb{4.5} & \bb{37.1} & \bb{24.8} & \bb{57.8} & \bb{14.2} & \bb{65.6} \\

\Xhline{0.3\arrayrulewidth}
\vitb{} & 81.6 & 25.9 & 47.0 & 33.3 & 70.2 & 19.8 & 49.1 \\
\gr + \lsd{} & \bb{82.3} & \bb{27.3} & \bb{48.3} & \bb{35.0} & \bb{71.2} & \bb{21.1} & \bb{47.4} \\

\end{tabular}
} 
\caption[caption]{\textbf{Robustness evaluation.} The accuracy gain achieved with our methods is consistent across various distributional shifts. }
\label{tab:robustness}
\end{table}

\clearpage

\section{Loss Landscape}

\begin{figure*}[t]
\vspace{1em}
  \centering
  \includegraphics[width=.46\linewidth]{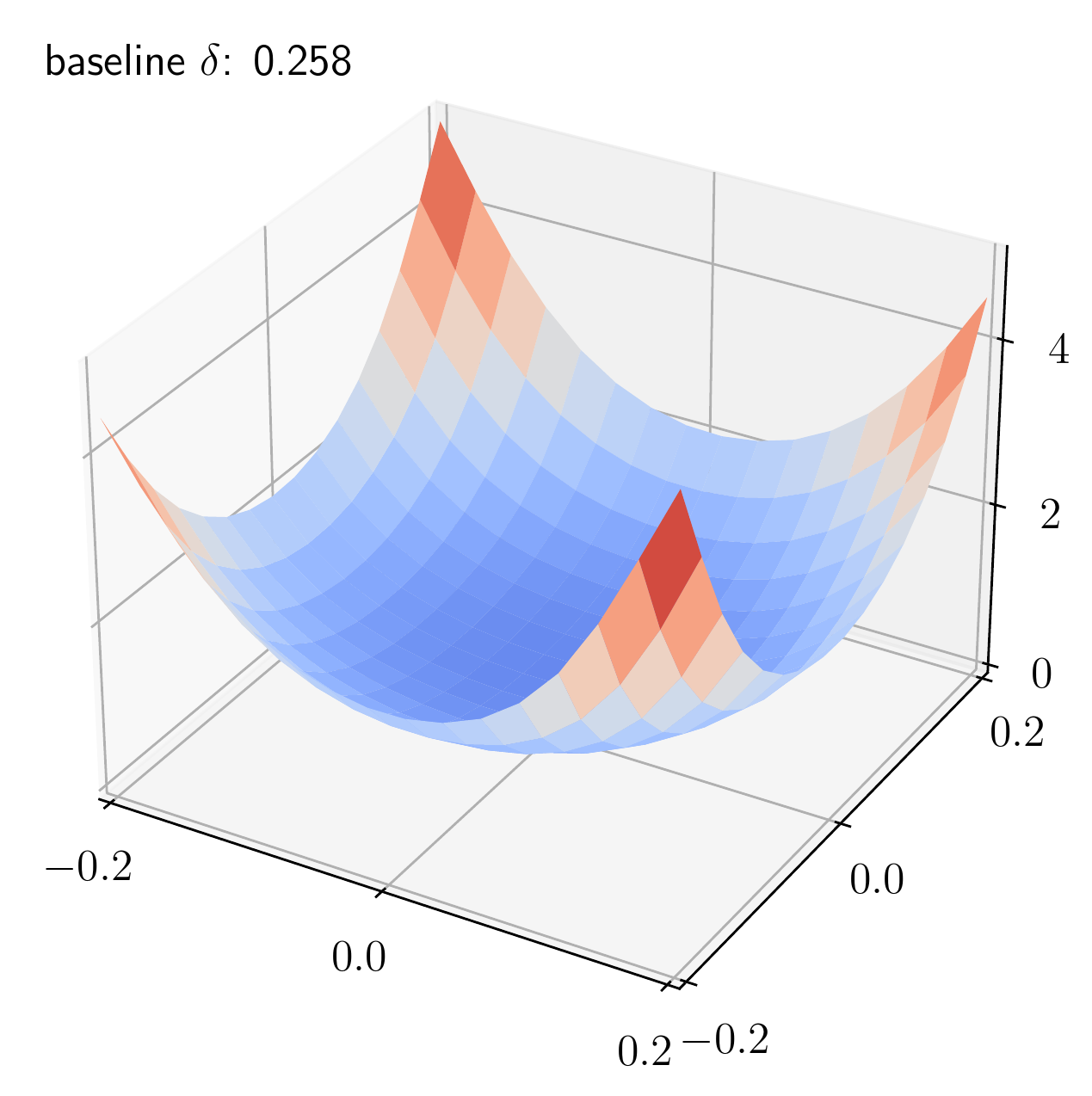}
  \hspace{-.3em}
  \includegraphics[width=.46\linewidth]{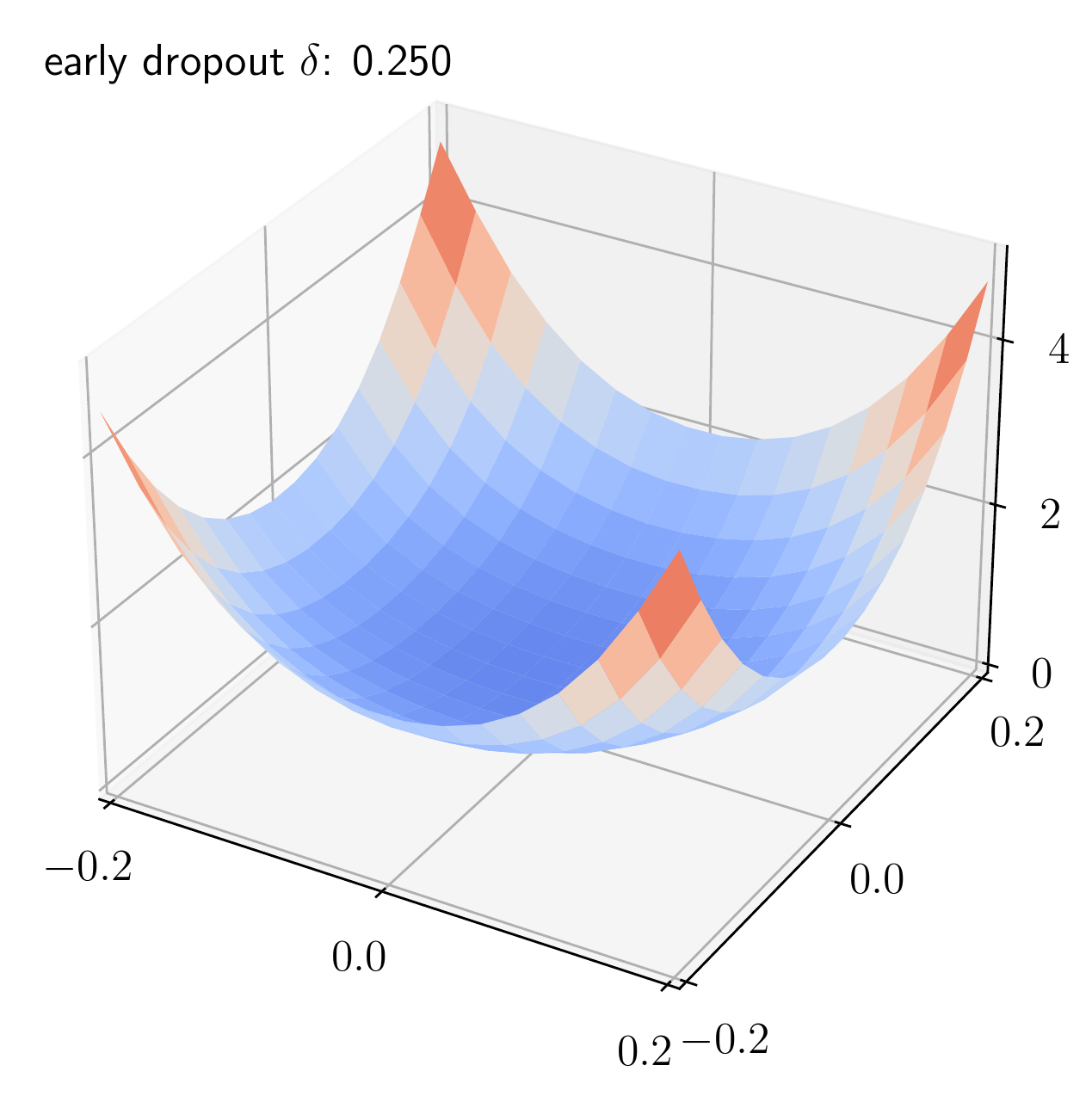}
  \vspace{-1em}
  % \vspace{.5em}
  \caption{\textbf{Loss Landscape Visualization}~\cite{li2018visualizing} for the baseline (left) and early dropout (right) models. Both models show similar levels of flatness both visually and when measured with the curvature metric $\delta$.}
  \label{fig:loss-landscape}
\end{figure*}

 We visualize the loss landscape~\cite{li2018} of \vit{} models trained with and without early dropout in Figure~\ref{fig:loss-landscape}. From the figure, we do not observe any significant difference in flatness around the solution area. To  quantitatively measure the curvature, we calculate $\delta$, the average difference in loss values between neighboring points:
  $$ \delta = \frac{1}{|N|} \sum_{(p_i, p_j) \in N} |L(p_i) - L(p_j)|$$
  where $N$ is the set of all neighboring pairs of points on the loss landscape, and $L(\cdot)$ denotes the loss value at a given point. Smaller $\delta$ indicates a flatter landscape. We notice a very slight difference in $\delta$, with 0.250 for early dropout and 0.258 for baseline. This suggests that early dropout may not improve generalization by finding flatter regions, unlike other methods such as~\citet{li2018visualizing} and~\citet{chen2021vision}.

% \clearpage

\section{Limitations}
We show that early and late dropout can benefit the training of small and large networks in a range of supervised visual recognition tasks. However, the application of deep learning extends far beyond this, and further research is needed to determine the impact of early and late dropout on other areas, such as self-supervised pre-training or natural language processing. It would also be valuable to explore the interplay between early / late dropout and other factors such as training duration or optimizer choice.

Another intriguing behavior that our current analysis cannot fully explain is shown in the training curves in Figure~\ref{fig:curve}. Early dropout does \emph{not} result in a lower training loss during the early dropout phase, even though it eventually leads to a lower final loss. This observation holds true even when evaluating the training loss with dropout turned off. Therefore, it appears that early dropout and gradient error reduction enhance optimization not by accelerating the process, but possibly by finding a better local optimum. This behavior warrants further study for a deeper understanding.

\section{Societal Impact}
The training and inference of deep neural networks can take an excessive amount of energy, especially in the large model and large data era. Our discovery on early dropout could spark more interest in developing training techniques for small models, which have far lower total energy usage and carbon emission than large models. 

It is also important to note that the benchmark datasets used in this study were primarily designed for research purposes, and may contain certain biases~\cite{de2019does} and not accurately reflect the real-world distributions. Further research is needed to address these biases and develop training techniques that are robust to real-world data variability.
\end{document}